\documentclass[logo,copyright]{nvidiatechreport}

\usepackage{hyperref}
\usepackage{url}

\usepackage{amsmath}
\usepackage{amsfonts}
\usepackage{graphicx}
\usepackage{booktabs}
\usepackage{amssymb}
\usepackage{mathtools}
\usepackage{multirow}
\usepackage{natbib}
\usepackage{placeins} % provides \FloatBarrier to keep floats from passing a point

\usepackage{xspace}
\usepackage{enumitem}

\usepackage{caption}
\newcommand{\method}{CANTO\xspace}
\newcommand{\fullmethod}{CAD-Native Transformer Operator\xspace}

\newcommand{\drivaer}[0]{DrivAerML}

\newcommand{\hilift}[0]{HiLiftAeroML}

\title{CANTO: CAD-Native Transformer Operator for AI-Aided Engineering}

\author{Daniel Leibovici, Nikola Borislavov Kovachki, Dawon Ahn, Ruben Ohana, Ira J. S. Shokar, Abouzar Ghasemi, Semih Akkurt, Rishikesh Ranade, Neil Ashton, Jan Kautz, Jean Kossaifi \\ 
NVIDIA}

\date{}

\begin{abstract}
Modern engineering systems, from automobiles to aircraft, are designed by using precise, continuous parametric computer-aided design (CAD) models. 
Evaluating design changes through numerical simulation requires meshing the continuous geometry, a computationally expensive and often brittle process that can require manual intervention and replaces the continuous representation with a discrete approximation.
Most neural surrogates accelerate the simulation, but inherit this representation gap by relying on meshes, point clouds, voxels, or other sampled approximations of geometry. 
We introduce \method, a transformer neural operator that maps directly from continuous CAD geometry to physical fields, without meshing the input geometry.
We develop a theoretical framework for learning operators from geometric manifolds to function spaces of physical fields, representing geometry through sequences of parametric patches.
\method instantiates this framework by directly tokenizing non-uniform rational B-spline (NURBS) patches from their control points, knot vectors, and weights, and predicts continuous surface and volume fields at arbitrary query locations.
We evaluate \method on four automotive and aircraft aerodynamics industry benchmarks: AhmedML, WindsorML, DrivAerML, and HiLiftAeroML.
\method achieves state-of-the-art accuracy on most evaluated surface and volume prediction tasks, including a 19.8\% reduction in surface-pressure relative $L_2$ error compared with AB-UPT on HiLiftAeroML.
Differentiability with respect to CAD parameters further enables gradient-based inverse design of designs.
On AhmedML, \method identifies designs with 4.4 to 20.4\% lower drag than the best dataset designs satisfying the same volume and lift constraints, with the improvements verified using the same CFD setup used to generate the original dataset.
\end{abstract}

\begin{document}

\maketitle

\abscontent

\section{Introduction}

Modern engineering design is an iterative loop: 
designers modify parametric computer-aided design (CAD) models,
and numerical simulations evaluate their performance.
In vehicle and aircraft aerodynamics, 
this simulation is computational fluid dynamics (CFD),
which solves the governing Partial Differential Equations (PDEs) of fluid flow 
to predict velocity, pressure, and wall-shear stress and combines them into integrated quantities such as drag and lift.
The geometry used by designers is exact and continuous: 
CAD models represent surfaces parametrically, 
often using Non-Uniform Rational B-Splines (NURBS).
A NURBS surface is defined by a compact set of control points, rational weights, and knot vectors, 
which together specify a smooth map 
from a low-dimensional parameter domain to physical space.

\begin{figure}[ht]
  \centering
  \includegraphics[width=0.9\linewidth]{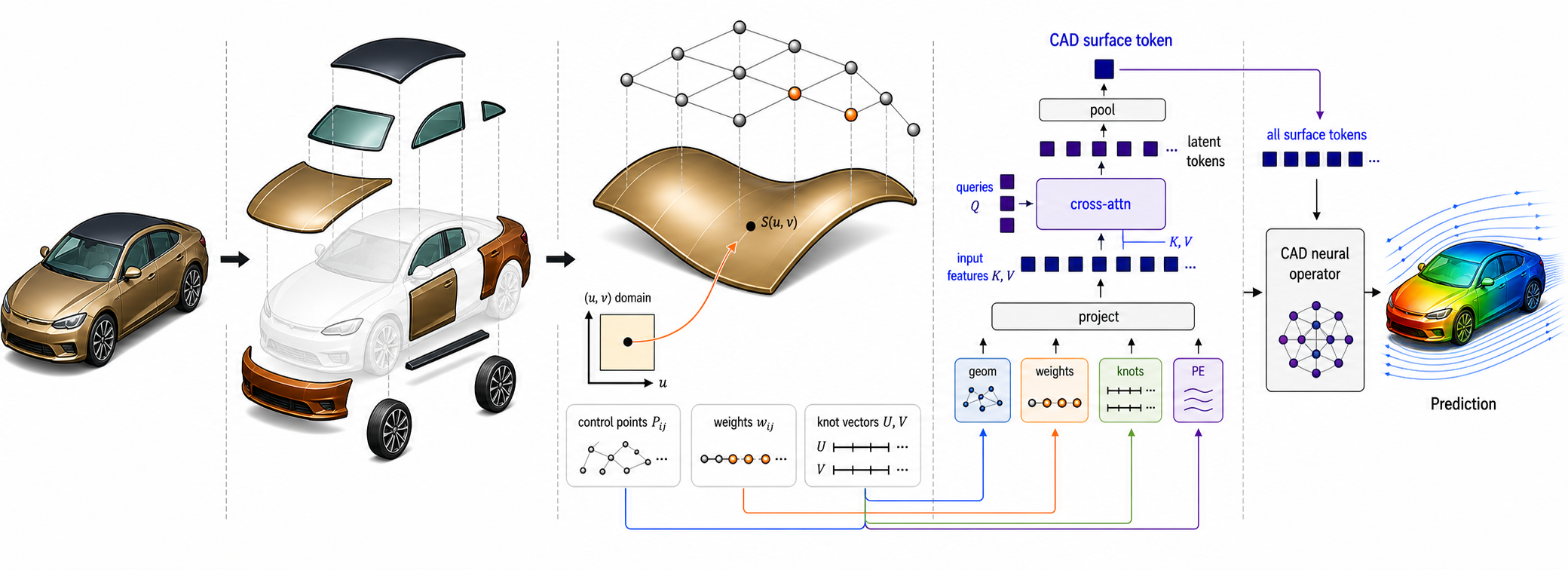}
  \caption{\textbf{Overview of our approach.}
  \method obviates the need for meshing step used by traditional AI surrogate pipelines by instead directly tokenizing continuous CAD NURBS geometries. 
  First, the original CAD geometry is expressed as a sequence of NURBS surfaces, before being encoded by their control points, knot vectors, and weights. These parametrized NURBS surfaces are then tokenized with a Perceiver-based encoder and used to condition our neural operator to predict the target physical fields.}
  \label{fig:teaser}% \vspace{-13pt}
\end{figure}

\textbf{The representation gap.}
CFD solvers do not typically operate directly on parametric CAD models.
Instead, the geometry is first discretized into a tessellated surface mesh
and a volumetric mesh, often comprising $10^7$ to $10^8$ cells, on which
the solver then solves the governing PDEs using finite-volume,
finite-element, or related numerical schemes. For industrial geometries, mesh generation is often the most brittle and labor-intensive stage of the analysis pipeline, requiring hours of manual cleaning, repair, and refinement. The mesh, not the CAD model, then becomes the
input representation for simulation and for most learned surrogates. We refer
to the gap between the exact CAD design and its sampled, mesh-based proxy as the
\emph{representation gap} between design and analysis. This gap keeps design and
analysis separated and bottlenecks the iterative engineering cycle.

\textbf{AI surrogates inherit the gap.}
Recent neural surrogates can accelerate CFD evaluation from days to milliseconds,
but they largely use the same kind of sampled geometric discretizations as classical solvers.
Graph neural operators and radius-graph models~\cite{li2020GNO,GINO},
point-cloud encoders~\citep{alkin2025ab,wu2024transolver,hao2023gnot}, and voxel-based networks all replace exact NURBS surfaces with a discretized mesh or point cloud before inference.
As a result, they accelerate analysis without closing the representation gap:
the model never sees the design representation used by the engineer, and its predictions remain conditioned on a discretized approximation of the geometry.
We discuss these methods and their resolution behavior in Section~\ref{sec:related}.

\textbf{Our approach.}
We propose the \fullmethod (\method), a transformer neural
operator that operates directly on the native parametric
NURBS representation of CAD geometry and predicts physical
fields as continuous functions queryable at arbitrary
spatial locations.
Instead of meshing, voxelizing, or spatially sampling the input
geometry, \method encodes each untrimmed NURBS surface into
a single latent token computed from its spline definition.
% We propose the \fullmethod (\method), a neural operator that operates directly on the native parametric NURBS representation of CAD geometry and predicts physical fields as continuous functions queryable at arbitrary spatial locations. Rather than meshing, voxelizing, or sampling the geometry, \method encodes each NURBS untrimmed surface into a single latent token computed from its spline definition.

A transformer then combines these patch tokens into a global geometry context, 
which conditions a transformer backbone 
on query points coordinate
for surface and volume field prediction, as well as engineering quantities.
This geometry encoding is independent of query resolution: it depends only on the
compact CAD description, instead of a discretized mesh approximation with millions of cells or points.
In summary, we make the following contributions: 
\begin{itemize}[leftmargin=*, itemsep=1pt, topsep=1pt, parsep=2pt, partopsep=4pt]
    \item \textbf{Framework.} We formalize CAD-to-physics surrogate modeling as a map from a class of compact embedded $C^1$-manifolds, represented through sequences of oriented boundary patch parametrizations, to a Banach function space of physical fields, providing a principled foundation for tokenizing parametric geometries.
    \item \textbf{Architecture.} We introduce a NURBS-native tokenization that constructs geometry and surface tokens directly from control points, weights, and knot vectors. Integrated into the transformer backbone of \method, these tokens enable continuous field prediction without mesh, voxel, or point-cloud-based discretization of the input geometry.
    \item \textbf{Empirical results.} On the AhmedML and WindsorML automotive aerodynamics
    benchmarks and the HiLiftAeroML aircraft aerodynamics benchmark, \method achieves
    state-of-the-art surface-field accuracy while decoupling input and output
    representations from CFD mesh resolution and fixed evaluation grids.
    By constructing surface context directly from parametric NURBS, \method
    retains accuracy as the spatial-context budget decreases by orders of
    magnitude, where point-cloud baselines collapse. On DrivaerML, it achieves
    competitive surface-field results and state-of-the-art volume-field accuracy.
    Finally, \method enables inverse design, identifying configurations with
    lower drag than any design in the original AhmedML dataset under the
    considered constraints.    
    \end{itemize}
The remainder of this paper is organized as follows: we review related works in~\ref{sec:related} and introduce relevant background material on NURBS in Section~\ref{sec:background}. We then introduce our proposed approach in Section~\ref{sec:method} and validate it empirically in Section~\ref{sec:experiments}.

\section{Related works}\label{sec:related}
\paragraph{Neural operators.}
Neural operators~\citep{kovachki2023neuraloperator}, such as the Fourier Neural
Operator~\citep{li2021fourier} and DeepONet~\citep{deeponet}, learn maps between
function spaces and have been applied broadly to systems governed by partial
differential equations (PDEs)~\citep{NO-nature, kovachki2024operator}. For PDEs on complex geometries,
however, the input domain is typically represented by a finite geometric proxy,
such as a mesh, point cloud, voxel grid, signed-distance field, or graph.
FNO and many of its variants operate on regular grids. Graph Neural
Operators~\citep{li2020GNO} extend neural operators to irregular domains by
aggregating information over neighborhoods defined by an $\epsilon$-radius
graph. GINO~\citep{GINO} uses GNO layers to map between point-cloud or
signed-distance representations and a regular latent grid, where an FNO can be
applied. These approaches improve geometric flexibility, but the geometry is
still supplied through a sampled approximation rather than the native CAD
representation.

\paragraph{Geometry-aware PDE and CFD surrogates.}
Recent work has extended neural operators and transformer-based surrogates to
irregular domains and large-scale engineering flows. GNOT~\citep{hao2023gnot},
OFormer~\citep{oformer}, Transolver~\citep{wu2024transolver},
GAOT~\citep{wen2025geometry}, and GINOT~\citep{liu2025geometry} use graph,
point-based, query-based, or attention mechanisms to model fields on complex
domains. DoMINO~\citep{domino}, AB-UPT~\citep{alkin2025ab},
FigConvNet~\citep{figconv}, MNO~\citep{wang2025mno}, and
Fusion-DeepONet~\citep{peyvan2025fusion} further demonstrate strong performance
on engineering and CFD benchmarks, including automotive aerodynamics and
high-speed flows.

These methods improve scalability and geometric flexibility, and
some can predict fields at arbitrary query locations.
However, their geometry encoders still operate on discretized geometric approximations: 
STL tessellations, mesh nodes, surface or volume point clouds, signed-distance samples, implicit grids, supernodes, or restricted parametric shape descriptions.
Thus, even when the output field is queried continuously, the input geometry has already been
discretized before reaching the model. 
In contrast, \method tokenizes the native CAD spline structure,
replacing mesh- or point-based geometry encoders
with NURBS patch tokens derived from its parametric
representation.

\paragraph{CAD representations for geometry learning.}
A growing body of work tries to use CAD models for geometric tasks. 
UV-Net~\citep{jayaraman2021uv} encodes boundary representations,
DeepCAD~\citep{wu2021deepcad} generates parametric CAD models,
NURBS-Diff~\citep{prasad2022nurbs} provides differentiable NURBS layers for surface fitting,
NeuroNURBS~\citep{fan2024neuronurbs} encodes NURBS surfaces directly from their parameters, and NURBSGen~\citep{usama2026nurbgen} fine-tunes Qwen3-4B~\citep{yang2025qwen3} using LoRA~\citep{hu2021lora} to map text descriptions to JSON sequences encoding NURBS surface parameters, which are subsequently converted into BRep geometry. These methods preserve CAD structure, but primarily target geometric representation learning, generation,
or reconstruction, rather than physical-field prediction.

Our approach is also related in spirit to isogeometric analysis (IGA)~\citep{hughes2005isogeometric}, which uses spline representations for both geometry and numerical analysis. However, IGA remains a numerical PDE discretization method: like traditional finite-element or finite-volume approaches, it solves the governing equations for each new configuration, using spline basis functions, typically NURBS, to represent the geometry and approximate the solution. In contrast, \method uses a parametric NURBS representation of the CAD geometry as the input space of a learned neural operator. Rather than performing a new numerical solve for each geometry, \method learns the solution operator from high-fidelity simulation data and directly predicts continuous physical fields and quantities of interest for unseen configurations at inference time.

\section{Preliminaries}\label{sec:background}

\paragraph{NURBS parametrizations.} The NURBS parameterization of curves and surfaces is based on B-spline basis functions constructed via the Cox--de Boor recursion~\citep{de1977package}. They are defined via a \emph{knot vector}, which is a non-decreasing sequence of real numbers
\[
  U = (u_0, u_1, \dots, u_{m+p+1}), \qquad u_k \le u_{k+1},
\]
where $p$ denotes the polynomial degree and $(m+1)$ the number of basis functions. Given a knot vector $U$, the B-spline basis functions $N_{k,p}^U$ are defined recursively, as described in Appendix~\ref{app:nurbs_param}. 
% For $p=0$,
% \[
%   N_{k,0}^{U} (u) =
%   \begin{cases}
%     1, & \text{if } u_k \le u < u_{k+1}, \\
%     0, & \text{otherwise},
%   \end{cases}
% \]
% and for $p \ge 1$,
% \[
%     N_{k,p}^U (u)
%   = \alpha_{k,p}(u) \, N^U_{k,p-1}(u)
%   + \beta_{k,p}(u) \, N^U_{k+1,p-1}(u),
% \]
% where
% \[
%     \alpha_{k,p}(u) = 
%     \begin{cases}
%         \frac{u - u_k}{u_{k+p} - u_k}, & u_{k+p} \neq u_k, \\
%         0, & \text{otherwise}
%     \end{cases}, \quad
%     \beta_{k,p}(u) = 
%     \begin{cases}
%         \frac{u_{k+p+1} - u}{u_{k+p+1} - u_{k+1}}, & u_{k+p+1} \neq u_{k+1}, \\
%         0, & \text{otherwise}.
%     \end{cases}
% \]
Given $n+1$ control points \(P_i \in \mathbb{R}^3\), for each \(0 \le i \le n\), and associated weights \(w_i \ge 0\), a \emph{NURBS curve} of degree \(p\) is defined as the mapping \(C : [0, 1] \to \mathbb{R}^3\),
\begin{equation*}
 C(u)
 =
 \frac{
     \sum_{i=0}^{n}
       N_{i,p}(u)\, w_i\, P_i
   }{
     \sum_{i=0}^{n}
       N_{i,p}(u)\, w_i
   }.
\end{equation*}

To define surfaces, we introduce an additional knot vector $V = (v_0, v_1, \dots, v_{s+q+1})$ where $v_l \leq v_{l+1}$ and its corresponding basis $N^V_{l,q}$. A surface can then be defined via the tensor product  $N^U_{k,p} \otimes N^V_{l,q}$. In particular, given a tensor-product grid of control points $P \in \mathbb{R}^{m \times s \times 3}$ and non-negative weights $w \in \mathbb{R}^{m \times s}$, a \emph{NURBS surface} of bi-degree \((p,q)\) is given by the mapping \(S : [0, 1] ^2 \to \mathbb{R}^3\),
\begin{equation}\label{eq:parametric_surface}
  S(u,v)
  =
  \frac{
      \sum_{k=0}^{m} \sum_{l=0}^{s}
        N_{k,p}^U(u)\, N_{l,q}^V(v)\, w_{kl}\, P_{kl}
    }{
      \sum_{k=0}^{m} \sum_{l=0}^{s}
        N_{k,p}^U(u)\, N_{l,q}^V(v)\, w_{kl}
    }.
\end{equation}
We will only consider NURBS surfaces that are non-intersecting, making $S$ injective.
% The unnormalized surface normal is given by the cross product of the partial derivatives,
% \begin{equation}
% \label{eq:surface_normal}
% \nu(u,v) = \frac{\partial S}{\partial u}(u,v) \times \frac{\partial S}{\partial v}(u,v).
% \end{equation}
% The orientation of the normal is induced by the parametrization of the surface, which in turn depends on the ordering of the control points $P_{kl}$ along the $u$- and $v$-directions. Reversing the indexing in either parametric direction (e.g., $k \mapsto m - k$ or $l \mapsto s - l$) flips the corresponding tangent direction and thus reverses the surface normal.
The unit surface normal is

\begin{equation}
\label{eq:surface_normal}
\nu(u,v)=
\frac{\partial_u S(u,v)\times\partial_v S(u,v)}
{\|\partial_u S(u,v)\times\partial_v S(u,v)\|},
\end{equation}

with orientation induced by the surface parametrization, and hence by the ordering of the control points $P_{kl}$ along the $u$- and $v$-directions. Reversing either parametric direction reverses $\nu$.

\paragraph{Trimmed NURBS surfaces.}
Trimming enables complex surface boundaries and holes to be represented while leaving the underlying NURBS parameterization unchanged. A \emph{trimmed NURBS surface} restricts the parametric domain of a NURBS surface using a set of trimming curves \(\mathcal{T} = \{C_1,\ldots,C_{N}\}\).
Specifically, while the underlying surface \(S(u,v)\) is defined over a $[0, 1] ^2$, the trimming curves in \(\mathcal{T}\), defined in the \((u,v)\)-parameter space, delimit a restricted parameter domain \(\mathcal{D} \subseteq [0,1]^2\).
The image of \(\mathcal{D}\) under \(S\) defines the corresponding \emph{patch} 
% \[
$
    \Sigma := S(\mathcal{D}) \subset \mathbb{R}^3,
$
% \] 
with restricted parameterization
% \[
$
\widetilde{S} := S|_{\mathcal{D}} : \mathcal{D} \rightarrow \Sigma.
$
% \]
% 
% 
We assume that \(S_{\mathcal{D}}\) is a diffeomorphism onto its image, so that each point on the patch $\Sigma$  is uniquely associated with a point in the trimmed parameter domain \(\mathcal{D}\) (which is the case for regular, non-self intersecting surfaces). In this work, we \textit{untrim} these surfaces by approximately decomposing them into collections of tensor-product NURBS patches, as detailed in Appendix~\ref{app:untrimming}. Specifically, each surface $\Sigma$ is decomposed as $\Sigma = \bigcup_{j=1}^{n_{\Sigma}} \Sigma_j$, where each patch $\Sigma_j$ is diffeomorphic to $[0,1]^2$.

% \paragraph{Untrimming NURBS Surfaces.}\label{trims} 

% Working with the trimming curves directly can be a practical challenge due to the complex nature of the way they specify the trimmed parameter domain $\mathcal{D}$. The work of~\citept{massarwi2018untrimming} proposes algorithms to untrim a NURBS surface by decomposing it into a collection of tensor-product NURBS patches. For a given patch $\Sigma$, this yields a decomposition $\Sigma = \bigcup_{j=1}^{n_{\Sigma}} \Sigma_{j},$
% where each $\Sigma_j$ is diffeomorphic to $[0,1]^2$ and represented by a NURBS parameterization $S_{j}$ given by \eqref{eq:parametric_surface}. We denote by $n_{\Sigma}$ the number of untrimmed surfaces and $S_{j}$ their respective maps for $k=1,\dots,n_{\Sigma}$. 
% These maps form the geometric inputs to our model (Section~\ref{tokenizing_manifolds}).

% While exact, the untrimming procedure of~\citept{massarwi2018untrimming} significantly increases the number of patches, the bi-degree of the NURBS surfaces, and the number of control points. 
% We instead use a boundary-first construction (Appendix~\ref{app:untrimming}) that exactly preserves watertightness, reproduces the input surface up to a small geometric residual (Table~\ref{tab:surface-error}), and keeps the number of patches, the bi-degree, and the number of control points comparable to those of the input.

\section{Methodology: CAD-Native Transformer Operator}\label{sec:method}

Herein, we introduce our framework for tokenizing manifolds and apply it to parametrize CAD geometries and condition a neural transformer operator resulting in our proposed method,~\method.

\subsection{A framework for manifold operator learning}

We present a general framework for the approximation of maps from a class of manifolds to a space of functions.  To that end, let $\Omega \subseteq \mathbb{R}^d$ be an open set and denote by $\mathcal{M}$ the set of finite-dimensional topological manifolds. Let $\mathcal{U}(\Omega;\mathbb{R}^k)$ be a Banach function space of $\mathbb{R}^k$-valued functions with domain $\Omega$. We consider the approximation of maps
\begin{equation*}
    \label{eq:general_map}
    F : \mathcal{M}' \subseteq \mathcal{M} \to \mathcal{U}(\Omega;\mathbb{R}^k),
\end{equation*}

where $\mathcal{M}'$ is a subclass of manifolds relevant to the specific problem. Such maps arise naturally in computational science whenever the solution of a physical system depends on its geometry, with examples including fluid dynamics, elasticity, and plasticity~\citep{li2020GNO, li2023fourier,GINO}. While our framework is general and can be applied to a variety of problems, we focus on a prototypical case arising in computational fluid dynamics, where the map $F$ represents the domain-to-solution map of a PDE and the manifold represents a solid obstacle within the computational domain.

To make this precise, suppose that $\Omega \subset \mathbb{R}^d$ is a bounded, Lipschitz domain. We will consider the set of pairs $(M,\varphi)$ where $M$ is a compact $d$-dimensional $C^1$-manifold with boundary and $\varphi: M \to \Omega$ is a $C^1$-embedding such that $\varphi(M) \subset \Omega$ is a compact \(d\)-dimensional embedded submanifold with boundary and $\Omega \setminus \varphi(M)$ is a bounded, Lipschitz domain. We then define $\mathcal{M}'$ to be the set of equivalence classes of such pairs, where two pairs are identified if their embeddings have the same image. In particular, the two manifold, embedding pairs $(M, \varphi)$ and $(N, \psi)$ are equivalent if there exists a homeomorphism $h : M \to N$ such  that $\varphi = \psi \circ h$.  We can therefore identify a point $(M,\varphi) \in \mathcal{M}'$ with the bounded Lipschitz domain $\Omega_{M,\varphi} \coloneqq \Omega \setminus \varphi(M)$. Given a PDE, we can then define $F$ as the domain-to-solution map extended to $\Omega$.

To illustrate this, we consider the Poisson equation, in particular,
\begin{align}
\begin{split}
\label{eq:poisson}
    - \Delta u &=f \qquad \text{in } \Omega_{M,\varphi}, \\
    u &= 0 \qquad \text{in } \partial \Omega_{M,\varphi},
\end{split}
\end{align}
for some fixed $f \in H^{-1}(\Omega)$. Standard elliptic theory implies that \eqref{eq:poisson} has a unique weak solution $u \in H^1_0 (\Omega_{M,\varphi})$ \citep{gilbarg2001elliptic}. Hence we can associate each point in $\mathcal{M}'$ with the PDE solution $u \in H^1_0 (\Omega_{M,\varphi})$. We can then identify each solution with its zero extension $\tilde{u}$ to all of $\Omega$ so that $\tilde{u}|_{\Omega_{M,\varphi}} = u$ and $\tilde{u}|_{\Omega \setminus \Omega_{M,\varphi}} = 0$. Clearly $\tilde{u} \in H^1_0(\Omega)$, so we define $F : \mathcal{M}' \to H_0^1(\Omega)$ by $(M,\varphi) \mapsto \tilde{u}$. PDEs with more complex boundary conditions can be treated with Sobolev extension operators \citep{stein1970singular}. In practice, we are usually only interested in $u$ on $\Omega_{M,\varphi}$ and do not directly approximate the extension.

Numerically, it is usually much more efficient to work with the boundary of the solid rather than the solid volume. To do this, we further restrict \(\mathcal{M}'\) to manifolds with boundary for which the oriented boundary determines a unique embedded solid up to the above equivalence relation. For each representative \((M,\varphi)\), we associate the boundary parametrization \(\varphi|_{\partial M} : \partial M \to \Omega\) together with the outward unit normal field \(\nu : \partial M \to \mathbb{S}^{d-1}\) of the embedded solid \(\varphi(M)\). Thus, on this restricted class, the equivalence class of \((M,\varphi)\) may be represented by the boundary data \((\partial M,\varphi|_{\partial M},\nu)\). The domain-to-solution map is unchanged, because this oriented boundary data determines the same embedded obstacle \(\varphi(M)\), and hence the same PDE domain \(\Omega\setminus \varphi(M)\).

\subsection{Tokenized manifolds by patch embeddings}
\label{tokenizing_manifolds}

To construct an approximation architecture, we represent the embedded boundary by a finite collection of local parametrizations. Rather than requiring a single global parametrization of the considered geometry, we assume that, for each admissible embedded solid \(\varphi(M)\), its boundary \(\Gamma = \partial \varphi(M)\) admits a finite patch representation
\[\Gamma = \bigcup_{j=1}^n \Gamma_j,\]
where each patch \(\Gamma_j\) is parametrized by a \(C^1\) map $\gamma_j : [0,1]^{d-1} \to \Gamma_j \subset \mathbb{R}^d$. 
Each \(\gamma_j\) is assumed to be a diffeomorphism from \((0,1)^{d-1}\) onto the interior of \(\Gamma_j\), with suitable compatibility conditions on overlapping patch boundaries. For each patch, we denote by $\nu_j : [0,1]^{d-1} \to \mathbb{S}^{d-1}$ the associated outward normal field and consider the geometric input as a finite sequence of functions $\bigl((\gamma_1,\nu_1),\ldots,(\gamma_n,\nu_n)\bigr)$
where $\gamma_j \in C^1([0,1]^{d-1};\mathbb{R}^d)$ and
$\nu_j \in C([0,1]^{d-1};\mathbb{S}^{d-1})$.
The domain-to-solution map can then be regarded as a map from a sequence space of boundary parametrizations to the function space of solutions,
\[F : \mathcal{L}  \subset 
\bigcup_{n \geq 1}
\left(
C^1([0,1]^{d-1};\mathbb{R}^d)
\times
C([0,1]^{d-1};\mathbb{S}^{d-1})
\right)^n
\to
\mathcal{U}(\Omega;\mathbb{R}^k),
\]
where $\mathcal{L}$ denotes the set of admissible sequences of boundary parametrizations. 

The domain-to-solution map is well-defined on \(\mathcal{M}'\) and passing to the patch parametrizations induces a pullback of this map to the sequence space \(\mathcal{L}\) which we continue to denote by $F$. This representation is not injective in that different patch decompositions, orderings, or parametrizations may describe the same embedded solid and must therefore be assigned the same output. This non-uniqueness is a general feature of parametrically representing geometric data. We address it by (i) representing each \(\gamma_j\) using coefficients in a prescribed basis, as described in Section~\ref{subsec:nurbs_parameterizaing}, and (ii) training with multiple patch decompositions of the same surface.

In the next subsection, we demonstrate a particular parameterization of each $\gamma_j$ from which each surface normal may be computed directly. In this way, our choice in defining $\gamma_j$ carries with it the orientation of the manifold. Let $\tilde{\gamma}_j$ denote these parameterizations. We then define our approximation architecture as a transformer $F_\theta : \mathcal{L}' \times \Omega \to \mathbb{R}^k$ trained to approximately satisfy
\begin{equation}
\label{eq:approx_arch}
F_\theta \big ( (\tilde{\gamma}_1,\dots,\tilde{\gamma}_n), x \big ) = F \big ( (\gamma_1,\nu_1),\dots, (\gamma_n,\nu_n) \big )(x) \qquad \forall \, x \in \Omega,
\end{equation}
where $\mathcal{L}'$ denotes the set of such parameterizations. The main mechanism in our architecture can be viewed as a generalization of the functional patched attention formulated in \citep{calvello2025continuum}. In the CAD setting considered in the next subsection, these representations are given directly by the NURBS descriptions of the surface patches.

\subsection{Parameterizing CAD Geometries}
\label{subsec:nurbs_parameterizaing}
% We now specialize the construction of
% Section~\ref{tokenizing_manifolds} to CAD geometries.
We focus on \(d=3\)
and consider CAD boundaries represented by trimmed NURBS surfaces. 
Each trimmed surface is
decomposed into a finite collection of untrimmed NURBS patches. We may
therefore write the boundary of the embedded solid as
\[
    \Gamma = \bigcup_{j=1}^{n} \Sigma_j,
\]
where each patch \(\Sigma_j\) is parameterized by a NURBS map
% \[
%     S_j : [0,1]^2 \rightarrow \Sigma_j.
% \]
$ S_j : [0,1]^2 \rightarrow \Sigma_j.$
This is the patch representation introduced in
Section~\ref{tokenizing_manifolds}, with the abstract patch maps
\(\gamma_j\) instantiated by
% \[
    $ \gamma_j = S_j. $
% \]

Furthermore, the outward normal field associated with each patch is determined
by its NURBS parameterization~\eqref{eq:surface_normal}, with the orientation chosen consistently with the outward orientation of
\(\Gamma\). Thus, both \(\gamma_j\) and \(\nu_j\) are determined by the NURBS
description of \(S_j\).

In particular, the finite-dimensional patch representation
% introduced in
% Section~\ref{tokenizing_manifolds} 
is given in the CAD setting by
\[
    \widetilde{\gamma}_j
    =
    \left(U^{(j)}, V^{(j)}, P^{(j)}, w^{(j)} \right),
\]
where $U^{(j)}$, $V^{(j)}$, $P^{(j)}$ and $w^{(j)}$ denote the
knot vectors, control points and weights defining \(S_j\).
Consequently, the sequence
% \[
$
    \left(
    \widetilde{\gamma}_1,\ldots,\widetilde{\gamma}_n
    \right)
$
% \]
provides the finite-dimensional representation of the CAD boundary supplied
to the approximation architecture.

\subsection{NURBS tokenization and operator architecture}
\label{method}

\method realizes the operator in
Eq.~\eqref{eq:approx_arch} by directly tokenizing the parametric NURBS
representation of CAD geometry. 
The resulting geometry context conditions a transformer
backbone that predicts surface and volume fields at
arbitrary query locations, without meshing the input geometry.

\paragraph{Tokenizing NURBS surfaces.}\label{tokenizing_geometries}
A key challenge in tokenizing CAD geometries is that each NURBS surface is parameterized by a control net and knot vectors whose sizes and configurations vary across surfaces.
Directly treating all NURBS parameters as geometry tokens would result in a variable and potentially large number of tokens per surface. 
We therefore propose an attention-based geometry encoder which, for each surface, takes the set of parameters in $\tilde{\gamma}_j$ and maps them to a fixed $d$-dimensional representation.

Applying this to all surfaces gives us a $\mathbb{R}^{n\times d}$ representation of the geometry, noting that $n$ can vary across geometries. This representation is the tokenized version of our input manifold as defined via the NURBS parameterization.

For the $j$-th surface, we first concatenate the control-point coordinates
$\mathbf{P}^{(j)} \in \mathbb{R}^{m_j \times s_j \times 3}$ with the
corresponding NURBS weights $\mathbf{w}^{(j)} \in
\mathbb{R}^{m_j \times s_j}$ to form the weighted control-point tensor
$\mathbf{P}^{(j)}_w \in \mathbb{R}^{m_j \times s_j \times 4}$.
Each control point is additionally augmented with its two-dimensional
control-net indices $(i,j)$, providing its position within the tensor-product
control grid. 

For  the two knot vectors in the $U$- and $V$-directions, we augment each knot value with its
position in the corresponding knot sequence,
% \[
% U^{(j)} \mapsto
% \left\{\left(U^{(j)}_k,\frac{1}{k}\right)\right\}_{k=1}^{m_j+p_j+2},
% \qquad
% V^{(j)} \mapsto
% \left\{\left(V^{(j)}_\ell,\frac{1}{\ell}\right)\right\}_{\ell=1}^{s_j+q_j+2},
% \]
\[
U^{(j)} \mapsto
\left\{\left(U^{(j)}_k, 1/k \right)\right\}_{k=1}^{m_j+p_j+1},
\qquad
V^{(j)} \mapsto
\left\{\left(V^{(j)}_\ell,1/\ell\right)\right\}_{\ell=1}^{s_j+q_j+1},
\]
using the inverse of the index to ensure the positional encoding is bounded. 

The resulting sequences are processed by three individual
Perceiver Pooling modules, which independently aggregate
information from the weighted control points and the two
knot vectors.
The modules share the same general architecture but use
component-specific hyperparameters, as detailed in
Appendix~\ref{appendix:model}.

Each module first projects the control-point
(resp. knot-vector) inputs to
$d^{\mathcal{P}}_{cp}$-dimensional
(resp. $d^{\mathcal{P}}_{kv}$-dimensional) embeddings,
which are combined with a sinusoidal positional encoding.
A Perceiver-style pooling block~\citep{jaegle2021perceiver}
then uses a fixed set of $q^{\mathcal{P}}_{cp}$
(resp. $q^{\mathcal{P}}_{kv}$) learned latent queries
to cross-attend to the input sequence, followed by
self-attention among the latent tokens.
This cross-attention/self-attention pooling block is
repeated for $K^{\mathcal{P}}_{cp}$
(resp. $K^{\mathcal{P}}_{kv}$) layers.
The three pooled representations of the weighted control
net and the two knot vectors are then concatenated,
and the resulting feature vector is projected through
an MLP to yield a fixed-dimensional patch token.

\paragraph{Global geometry context.}
The $n$ patch tokens are jointly processed by $K_g$
DiT-style self-attention blocks~\citep{peebles2023scalable},
allowing each token to incorporate global context from
the complete CAD geometry.
The resulting geometry tokens form an
$\mathbb{R}^{n\times d}$ representation, where $n$ can
vary across geometries.

\paragraph{Transformer backbone.}
For the solution branch of \method, we follow the
Anchored-Branched Universal Physics Transformer
(AB-UPT)~\citep{alkin2025ab}, a Transformer-based multi-branch
neural operator with separate branches for geometry,
surface, and volume data.
The NURBS-derived geometry tokens condition the surface
and volume streams through cross-attention.
We replace the spatially sampled surface anchors with
surface tokens obtained directly from the NURBS representation
(Appendix~\ref{appendix:anchor_split}) and encoded using
the above Perceiver Embedding procedure.
Volume anchors remain randomly sampled from the volume coordinate
set, as in the standard AB-UPT. We use the same anchored neural field decoder, which predicts fields at arbitrary query locations without quadratic attention over query points. Further architectural details are provided in Appendix~\ref{appendix:model}.

\section{Experiments}\label{sec:experiments}
% CAN BE REMOVED IF NEED SPACE
% We evaluate the performance of \method, by comparing it with existing methods and thorough ablations on several industry-standard benchmark datasets for automotive and non-automotive computational fluid dynamics. 

% \paragraph{Experimental setting}
We evaluate \method against state-of-the-art methods on three industry standard benchmark datasets for automotive computational fluid dynamics: AhmedML~\citep{ashton2024ahmedml}, WindsorML~\citep{ashton2024windsorml} and DrivaerML~\citep{ashton2024drivaerml}.
We further validate the performance of \method on a recent, non-automotive dataset, HiLiftAeroML~\citep{ashton2026hiliftaeromlhighfidelitycomputationalfluid}. It contains \(1,800\) high-lift aircraft geometries and large-scale CFD simulations with approximately 200 million volume cells and 30 million surface cells per sample.
We expand on each dataset in the Appendix~\ref{appendix:dataset}.
Details regarding model architectures and training are provided in Section~\ref{appendix:model}.

\subsection{Performance}
% First, we compare the performance of our approach against existing works. 
Table~\ref{tab:perform} compares \method with state-of-the-art methods across four benchmark datasets. In particular, we trained AB-UPT on WindsorML and HiLiftAeroML, while the remaining baseline results are taken from~\citep{alkin2025ab}.
On AhmedML dataset, \method achieves the lowest error on three of four fields, and , especially, reduces the surface pressure error $\boldsymbol{p}_s$ of the state-of-the-art method by 11.6\%.
On \drivaer, \method is on par with AB-UPT~\citep{alkin2025ab}; it achieves the lowest volume velocity error $\boldsymbol{u}$ while AB-UPT marginally remains better for the surface pressure. Finally, on the \hilift\,  dataset, CANTO achieves the lowest error on all considered fields:  it reduces the relative $L_2$ error by 19.8\% for surface pressure, 2.3\% for wall-shear stress and 16.9\% for velocity. 
\method obtains nearly perfect accuracy in drag coefficients, with an $R^2$ of $0.99$, improving over AB-UPT ($0.96$).
On WindsorML dataset, \method attains the lowest error on every field, lowering the surface pressure error of AB-UPT by 5.5\%.

% In application, integrated quantities are often the most relevant for engineers. \method achieves superior performance for drag, with a relative error of $9\%$ (lower is better) and an $R^2$ of $0.65$ (higher is better), while AB-UPT gets $10.4\%$ and $0.47$, respectively.
% 

\begin{table}[h]
\centering
\caption{\textbf{Comparison of \method with prior methods.} We report the relative L2 errors (\%) for the surface pressure $\boldsymbol{p}_s$, volume velocity $\boldsymbol{u}$, volume vorticity $\boldsymbol{\omega}$, and wall shear stress $\boldsymbol{\tau}$ fields on the AhmedML, DrivAerML, WindsorML, and HiLiftAeroML datasets.}
\label{tab:perform}

\setlength{\tabcolsep}{3.5pt}

\begin{tabular}{
@{}l@{\hspace{1pt}}
cccc
@{\hspace{5pt}}ccc
@{\hspace{5pt}}ccc
@{\hspace{5pt}}ccc@{}
}
\toprule
& \multicolumn{4}{c}{\textbf{AhmedML}}
& \multicolumn{3}{c}{\textbf{DrivAerML}}
& \multicolumn{3}{c}{\textbf{WindsorML}}
& \multicolumn{3}{c}{\textbf{HiLiftAeroML}} \\
\cmidrule(lr){2-5}
\cmidrule(lr){6-8}
\cmidrule(lr){9-11}
\cmidrule(lr){12-14}

\textbf{Method}
& $\boldsymbol{p}_s$ & $\boldsymbol{u}$ & $\boldsymbol{\omega}$ & $\boldsymbol{\tau}$
& $\boldsymbol{p}_s$ & $\boldsymbol{u}$ & $\boldsymbol{\tau}$
& $\boldsymbol{p}_s$ & $\boldsymbol{u}$ & $\boldsymbol{\tau}$
& $\boldsymbol{p}_s$ & $\boldsymbol{u}$ & $\boldsymbol{\tau}$ \\
\midrule

PointNet
& 8.02 & 5.44 & 66.04 & 10.09
& 23.63 & 28.13 & 41.85
& -- & -- & --
& -- & -- & -- \\

Graph U-Net
& 6.46 & 4.15 & 53.66 & 7.29
& 16.13 & 17.98 & 27.84
& -- & -- & --
& -- & -- & -- \\

GINO
& 7.90 & 6.23 & 71.81 & 8.18
& 13.03 & 40.58 & 21.71
& -- & -- & --
& -- & -- & -- \\

LNO
& 12.95 & 7.59 & 72.49 & 11.50
& 20.51 & 23.27 & 36.44
& -- & -- & --
& -- & -- & -- \\

UPT
& 4.25 & 2.73 & 15.03 & 5.80
& 7.44 & 8.74 & 12.93
& -- & -- & --
& -- & -- & -- \\

OFormer
& 4.12 & 3.63 & 15.06 & 4.60
& 4.85 & 6.64 & 8.92
& -- & -- & --
& -- & -- & -- \\

Transolver
& 3.45 & 2.05 & 8.22 & 4.00
& 4.81 & 6.78 & 8.95
& -- & -- & --
& -- & -- & -- \\

Transformer
& 3.41 & 2.09 & 6.76 & 4.03
& 4.35 & 6.21 & 8.26
& -- & -- & --
& -- & -- & -- \\

AB-UPT
& \underline{3.01} & \textbf{1.90} & \underline{6.52} & \underline{3.88}
& \textbf{3.82} & \underline{5.93} & \underline{7.29}
& \underline{7.21} & \underline{7.04} & \underline{5.65}
& {5.65} & \underline{7.06}  & \underline{8.72}  \\

\midrule

\textbf{\method} % CANTO-GA
& \textbf{2.66} & \underline{2.04} & \textbf{4.83} & \textbf{3.53}
& \underline{3.97} & \textbf{5.14} & \textbf{7.15}
& \textbf{6.81}  & \textbf{6.69} & \textbf{5.46}
& \textbf{4.53}  & \textbf{5.87} & \textbf{8.52} \\

\bottomrule
\end{tabular}
\end{table}
The inference latency, peak memory requirements, and an end-to-end runtime comparison of our method and the traditional CFD framework used to generate these datasets are presented in Appendices~\ref{appendix:latency}--~\ref{pipeline_comparison}.

\subsection{Ablation Study}

\paragraph{Fixed-size vs. Variable-size NURBS features.} 

First, we ablate the geometry encoder of \method with fixed-size and variable-size NURBS features. In the fixed-size variant, the encoder is a simple MLP, since the network sees the same shape of inputs. In the variable-size variant, a perceiver-based encoder is employed to handle the varying sizes of control points and knot vectors. Table~\ref{tab:ahmedml:abl} compares the error of these variants. They achieve comparable error across all four fields; the fixed-size variant performs slightly better on the surface pressure ($\boldsymbol{p}_s$ , while the variable-size variant performs better on the wall-shear stress $\boldsymbol{\tau}$) and on the volume fields ($\boldsymbol{u}$ and $\boldsymbol{\omega}$). While the fixed-size variant is simpler, the variable-size variant is practically preferred since real-world CAD surfaces are highly heterogeneous.

\begin{table}[h]
    % \vspace{-5mm}
    \centering
    \caption{
    \textbf{Evaluation of \method with fixed-size and variable-size NURBS features} on the AhmedML dataset.
    }
    \label{tab:ahmedml:abl}
    \setlength{\tabcolsep}{4pt}
    \begin{tabular}{l cccc cccc ccc}
    \toprule
    \textbf{Method} & $\boldsymbol{p}_s$  & $\boldsymbol{u}$  & $\boldsymbol{\omega}$ & $\boldsymbol{\tau}$ \\
    \midrule
    Fixed-size& \textbf{2.50} & {2.27} & {5.80} & 3.54 \\
    Variable-size & 2.66 & \textbf{2.04} & \textbf{4.83} & \textbf{3.53} \\
    \bottomrule
    \end{tabular}
    % \end{table}
\end{table}

\paragraph{Robustness to sampling.}
\begin{figure}[htb!]
  \centering
  \includegraphics[width=\linewidth]{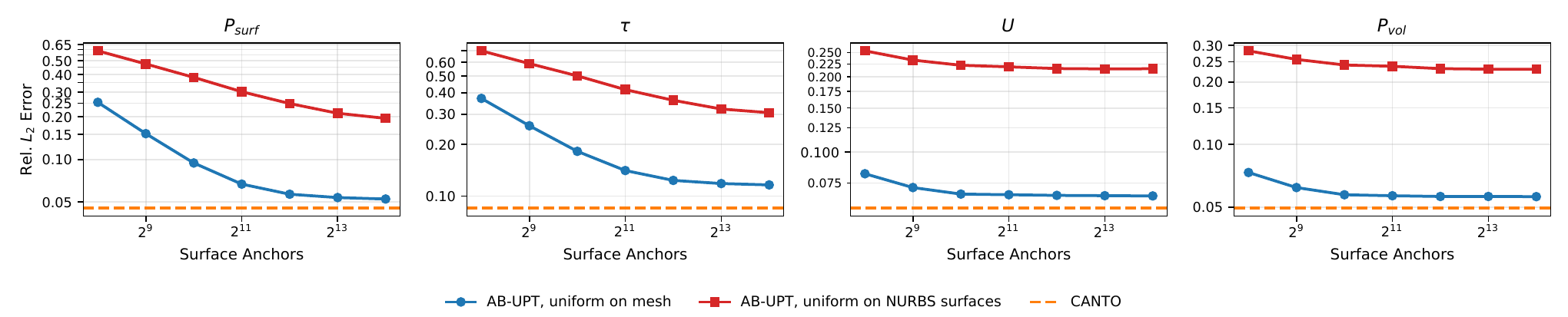}
  \caption{
  \textbf{Robustness to sampling procedure on the HiLiftAeroML dataset.} Here, \emph{uniform-on-mesh} denotes anchors sampled uniformly from the surface mesh points, as in AB-UPT, while \emph{uniform-on-NURBS surfaces} samples anchors uniformly in $(u,v)$ parameter space on each NURBS surface, with the number of anchors allocated to each surface proportional to its area.
  }
  \label{fig:sampling_robustness}
\end{figure}
Figure~\ref{fig:sampling_robustness} highlights a key advantage of CANTO: its geometry tokens are derived directly from the NURBS representation and are therefore independent of mesh resolution, anchor count, and sampling strategy. In contrast, AB-UPT exhibits strong sensitivity to its surface-anchor distribution, with different sampling strategies converging to distinct error levels even at large anchor budgets. CANTO avoids this sampling dependency entirely and achieves lower errors across all tested anchor budgets and quantities.

\subsection{CANTO-based inverse design}\label{sec:opt-task}

By operating directly on the parametric CAD representation, CANTO is differentiable with respect to the underlying geometry parameters, enabling gradient-based optimization for engineering design objectives. In automotive aerodynamics, for instance, reducing drag while controlling lift or downforce is a central design objective, typically subject to geometric and other engineering constraints. In what follows, we demonstrate how CANTO can be used to formulate and efficiently solve such a constrained shape-optimization problem directly in the CAD parameter space.

\begin{figure}[tbh!]
  \vspace{-2mm}
  \centering
  \vspace{-0.5\baselineskip}
  \includegraphics[width=\linewidth]{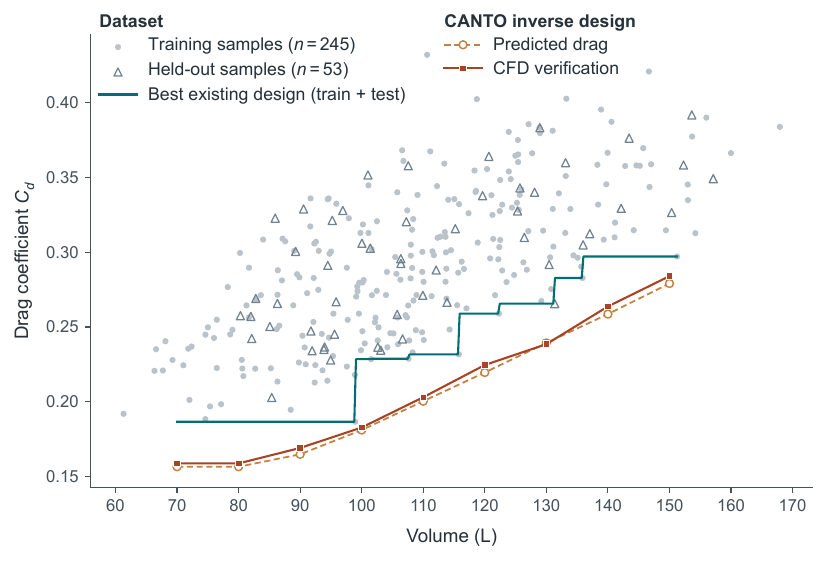}
  \caption{\textbf{Optimized Pareto front via inverse design.}
  Grey and blue markers show the ground-truth drag of training and held-out
  designs satisfying $C_l \leq 0.05$, while the blue staircase denotes the
  best dataset design satisfying each volume floor $V \geq V_0$. The dashed
  and solid red curves show the CANTO-predicted optima and their CFD
  validations, respectively. At every volume floor, CANTO identifies valid
  designs beyond the dataset benchmark front, with predictions closely
  matching CFD.}
  \label{fig:sweep_front_validated}
  \vspace{-0.5\baselineskip}
\end{figure}

\paragraph{Problem statement.}
We minimize aerodynamic drag over the six-parameter AhmedML body family, namely, \textit{body length}, \textit{body height}, \textit{body width}, \textit{front arc diameter}, \textit{slant angle length} and \textit{slant angle height}, subject to a lower bound on enclosed volume and an upper bound on lift,
\begin{equation}
\label{eq:problem}
\min_{x \in \mathcal{X}} \; D(x)
\quad \text{s.t.} \quad
V(x) \geq V_0, \qquad
L(x) \leq \varepsilon ,
\end{equation}
where $x$ collects the design parameters, $\mathcal{X}$ is the axis-aligned box
spanned by the dataset described in Table~\ref{tab:design-space}, and $\varepsilon = 0.05$.

\paragraph{CANTO-based optimization.}
$D$ and $L$ are predicted by CANTO, trained exclusively on the AhmedML training split, with the remaining validation and test cases held out from training. Because both the mapping from the design variables $\mathbf{z}$ to the NURBS geometry and the CANTO surrogate are differentiable, the gradients $\nabla_{\mathbf{z}} D$, $\nabla_{\mathbf{z}} V$, and $\nabla_{\mathbf{z}} L$ can be obtained directly through automatic differentiation. We leverage these gradients to solve~\eqref{eq:problem} using SLSQP~\citep{kraft1994algorithm}, supplying them explicitly to the optimizer. Starting from the lowest-drag feasible design in the entire dataset, the optimization then explores the continuous design space using the surrogate-derived gradient information.

\paragraph{Results.}
The optimized designs achieve improvements across the entire front, with gains ranging from $4.4\%$ to $20.4\%$ relative to the best dataset designs at the corresponding volume thresholds, as shown in Figure~\ref{fig:sweep_front_validated}. We further validate the optimized candidates using a traditional CFD solver (see Appendix~\ref{Optimization}), which confirms the predicted performance gains. Across these designs, CANTO's predictions remain within $2.2\%$ of the corresponding CFD values, demonstrating the surrogate's accuracy even for geometries discovered through optimization and not present in the original dataset. The CFD simulations also confirm that the lift constraint is satisfied at every volume threshold. Together, these results demonstrate that CANTO can successfully leverage its differentiable CAD representation to perform constrained design optimization and identify improved designs beyond those available in the original dataset.

% \begin{figure}[h!]
%   \centering
%   \includegraphics[width=\linewidth]{Figures/inverse_design.pdf}
%   \caption{\textbf{Optimized Pareto front via inverse design.}  Grey and blue markers show the ground-truth drag of training and held-out designs satisfying $C_l \leq 0.05$, while the blue staircase denotes the best dataset design satisfying each volume floor $V \geq V_0$. The dashed and solid red curves show the CANTO-predicted optima and their CFD validations, respectively. At every volume floor, CANTO identifies valid designs beyond the dataset benchmark front, with predictions closely matching CFD.
%   }
%   \label{fig:sweep_front_validated}
% \end{figure}

\FloatBarrier
\section{Conclusion}
Our proposed \fullmethod closes the representation gap between CAD and learned physics.
It takes parametric CAD directly as input to a neural transformer operator, 
without first reducing it to a mesh, point cloud, or implicit field. 
On AhmedML, WindsorML, DrivaerML and HiLiftAeroML, it reaches state-of-the-art accuracy on most evaluated surface-field and volume-field prediction tasks, with a 19.8\% relative reduction in surface-pressure error on HiLiftAeroML. 
By learning directly from NURBS representations, \method eliminates
the need for meshing at inference and remains robust to changes
in discretization and sampling.
\method is differentiable with respect to its NURBS inputs, 
enabling gradient-based shape optimization and inverse design 
directly in the CAD parameter space.

\bibliographystyle{unsrt}
\bibliography{references}

\appendix

\section{Model designs}
\label{appendix:model}

\begin{figure}[htb]
  \centering
  \includegraphics[width=0.8\linewidth]
    {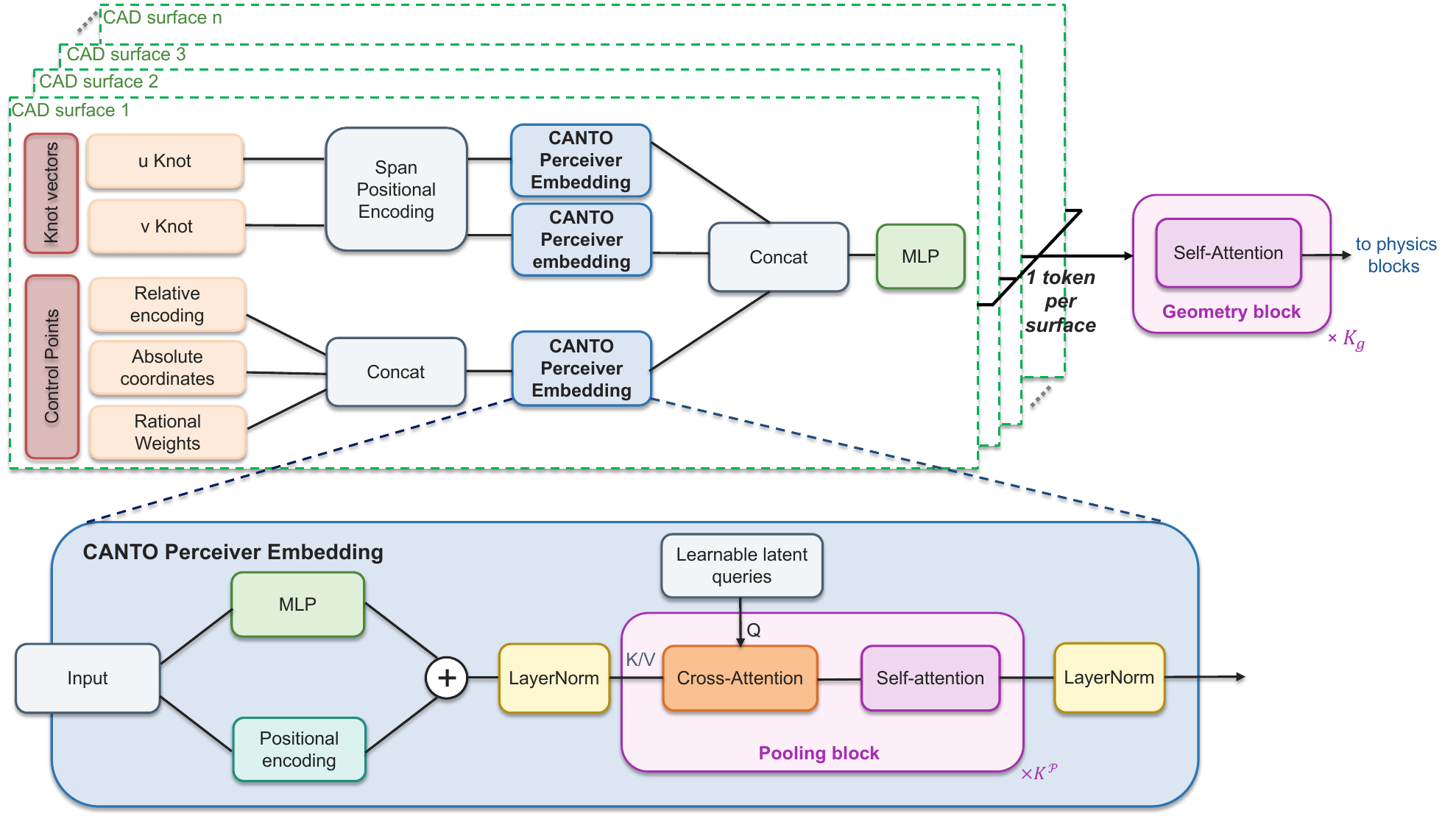}
  \caption{
  \textbf{CANTO geometry encoder architecture.}
  NURBS surfaces are represented by their control points,
  weights, and knot vectors.
  Three Perceiver Pooling modules independently encode
  the weighted control points and the two knot vectors.
  Their outputs are concatenated and projected to produce
  one fixed-dimensional token per patch.
  Self-attention across patch tokens incorporates global
  context from the complete CAD geometry.
  }
  \label{fig:encoder}
  \vspace{10pt}
\end{figure}

Here, we detail the transformer backbone and training
configuration used by \method.
Figure~\ref{fig:encoder} illustrates the NURBS geometry
encoder described in Section~\ref{tokenizing_geometries}.

\subsection{Backbone architecture}

For the solution branch of \method, we follow the
Anchored-Branched Universal Physics Transformer
(AB-UPT)~\cite{alkin2025ab}, a Transformer-based multi-branch
neural operator featuring dedicated streams for geometry,
surface, and volume data.
While AB-UPT relies on a point-cloud geometry branch,
\method replaces this with the CAD geometry encoder
described in Section~\ref{tokenizing_geometries} and
illustrated in Figure~\ref{fig:encoder}.
The resulting geometry tokens act as the geometry
key-value cache for the cross-attention blocks,
conditioning the surface and volume streams.

In AB-UPT, surface anchors are sampled from a point cloud.
In \method, these are replaced by surface tokens obtained
from a refined NURBS decomposition, as described in
Appendix~\ref{appendix:anchor_split}.
Each refined patch is encoded using the Perceiver
Embedding procedure described in
Section~\ref{tokenizing_geometries}.
The surface tokens provide a geometry-derived context set
for the surface stream.
During training and inference, arbitrary surface locations
are supplied as query tokens decoded against this context.
Volume anchors remain sampled from the volume coordinate
set, as in the standard AB-UPT.

The key property we inherit is the anchored neural field
decoder, which allows the model to predict fields on dense
domains without applying quadratic attention to all query
points.
Let $X^s = \{\mathbf{x}^s_i\}_{i=1}^{N_s}$ and $X^v = \{\mathbf{x}^v_i\}_{i=1}^{N_v}$ denote the surface and volume coordinates. These coordinates specify the query locations at which
surface and volume fields are predicted.

The first stage injects the NURBS-based latent geometry tokens $\mathbf{G}^{(L_{\mathrm{geo}})}$ into the surface and volume streams via cross-attention ($\operatorname{CA}$):
\begin{equation}
  H_s^{(0)} = \operatorname{CA}\left(E_s(X^s), \mathbf{G}^{(L_{\mathrm{geo}})}, \mathbf{G}^{(L_{\mathrm{geo}})}\right), \quad
  H_v^{(0)} = \operatorname{CA}\left(E_v(X^v), \mathbf{G}^{(L_{\mathrm{geo}})}, \mathbf{G}^{(L_{\mathrm{geo}})}\right),
\end{equation}
where $E_s$ and $E_v$ encode the coordinates. This conditions the predicted physical fields at every prediction location on the exact global shape.

Tokens then pass through a stack of shared physics blocks. Each block applies intra-branch self-attention ($\operatorname{SA}$) followed by inter-branch cross-attention:
\begin{align}
  \tilde{H}_{s,v}^{(\ell)} &= \operatorname{SA}_{\ell}\left(H_{s,v}^{(\ell-1)}\right), \\
  H_s^{(\ell)} &= \operatorname{CA}_{\ell}\left(\tilde{H}_s^{(\ell)}, \tilde{H}_v^{(\ell)}, \tilde{H}_v^{(\ell)}\right), \quad
  H_v^{(\ell)} = \operatorname{CA}_{\ell}\left(\tilde{H}_v^{(\ell)}, \tilde{H}_s^{(\ell)}, \tilde{H}_s^{(\ell)}\right). \nonumber
\end{align}
Sharing weights enforces a consistent physical update rule across both streams. These layers use standard pre-normalized Transformer blocks with MLP sublayers and Rotary Position Embeddings (RoPE). Finally, independent decoders and linear heads predict fields like pressure, velocity, and wall shear stress.

To avoid $O(N^2)$ complexity for large $N$, AB-UPT uses a small subset of $M \ll N$ anchor tokens $H_A$. %Attention for a token set $H$ is computed as:
%\begin{equation}
%  \operatorname{AnchorAttn}(H;H_A) = \operatorname{softmax}\left(\frac{Q(H) K(H_A)^\top}{\sqrt{d}}\right) V(H_A).
%\end{equation}
The anchors attend to one another and serve as a context set for decoding query tokens. This limits the quadratic bottleneck to $O(M^2)$, while decoding $N$ queries scales linearly, enabling practical million-scale CFD predictions. 
In \method, NURBS-derived surface tokens provide the
context for surface prediction, while volume anchors
remain sampled from the volume coordinate set.

\subsection{Hyper-parameters and Training} \label{appendix:hyperparameters}

\paragraph{Training objective}
Following AB-UPT, we train the complete model end to end using
$$ \mathcal{L} = 
\operatorname{MSE}(\mathbf{y}_s,\hat{\mathbf{y}}_s) + \lambda\operatorname{MSE}(\mathbf{y}_v,\hat{\mathbf{y}}_v), $$

where
$$ \hat{\mathbf{y}}_s = 
D^s\!\left(F^s\left(X^s,E^g(X^g),Z^v\right)\right), \qquad \hat{\mathbf{y}}_v =
D^v\!\left(F^v\left(X^v,E^g(X^g),Z^s\right)\right). $$

Here, $X^s$, $X^v$, and $X^g$ denote the surface queries, volume queries, and parameterized NURBS geometry, respectively. The normalized targets are $\mathbf{y}_s=(p_s,\tau_x,\tau_y,\tau_z)$ and $\mathbf{y}_v=(p_v,u_x,u_y,u_z)$. The operators $E$, $F$, and $D$ denote the geometry encoder, physics blocks, and field decoders, while $Z^s$ and $Z^v$ are the corresponding surface and volume latent representations. The coefficient $\lambda$ balances the surface and volume objectives.

% Architecture of the Geometry Encoder
\paragraph{Hyper-parameter validation}
For our AB-UPT backbone, we follow the same parameters and settings as the authors~\cite{alkin2025ab}, which are enumerated in Table~\ref{tab:phys_blocks_config}. Note that in each case we optimize the hyper-parameters (e.g., learning rate) over a small validation set. We train our model with Lion optimizer~\cite{chen2023symbolic} with a cosine learning rate schedule with a linear warmup. The exact configuration and hyper-paramterers used for each dataset are provided in Tables~\ref{tab:model_config} and~\ref{tab:training_config}.

\begin{table}[tb]
\centering
\caption{Geometry encoder hyper-parameters.}
\label{tab:model_config}
\begin{tabular}{l c c c c c }
\toprule
\textbf{Parameter} & \textbf{AhmedML} & \textbf{WindsorML} & \textbf{DrivaerML} & \textbf{HiLiftAeroML}\\
\midrule
  $q^{\mathcal{P}}_{cp}$ &   $8$  & $16$ & $4$ & $16$ \\
  $d^{\mathcal{P}}_{cp}$  &  $128$ & $192$ & $128$ & $192$   \\
  $K^{\mathcal{P}}_{cp}$  &  $4$  & $2$ & $4$ & $2$ \\  
  $q^{\mathcal{P}}_{kv}$  &  $4$ & $4$ & $4$ & $4$ \\
  $d^{\mathcal{P}}_{kv}$  &  $64$  & $64$ & $64$ & $64$ \\
  $K^{\mathcal{P}}_{kv}$  &  $1$ & $2$ & $1$ & $2$ \\
  Perceiver block attention heads &  $4$ & $2$ & $4$ & $2$ \\
  $d$  &  $192$ & $192$  & $192$ & $192$\\
  $K_g$ &  $6$  & $6$ & $18$ & $6$  \\
  Geom. block attention heads & $3$ & $3$ & $3$ & $3$  \\
\bottomrule
\end{tabular}
\end{table}

\begin{table}[t]
\centering
\caption{\textbf{Physics and decoder blocks hyper-parameters.}}
\label{tab:phys_blocks_config}
\begin{tabular}{l c}
\toprule
\textbf{Parameter} &  \\
\midrule
  Hidden dim.              & $192$ \\
  Attention heads          & $3$   \\
  Surface decoder depth          & $6$   \\
  Volume decoder depth          & $6$   \\
\bottomrule
\end{tabular}
\end{table}

\begin{table}[t]
\centering
\caption{\textbf{Training hyper-parameters.}}
\label{tab:training_config}
\begin{tabular}{l c c c c}
\toprule
\textbf{Parameter} & \textbf{AhmedML} & \textbf{DrivAerML} & \textbf{WindsorML} & \textbf{HiLiftAeroML} \\
\midrule
  Learning rate            & $5{\times}10^{-5}$ & $1{\times}10^{-4}$ & $3{\times}10^{-4}$ & $2{\times}10^{-4}$ \\
  Weight decay             & $0.05$ & $0.05$ & $0.05$ & $0.05$ \\
  % & Training epochs          & 2000 & 500 & 2000 \\
  Training steps           & $200\text{k}$ & $200\text{k}$ & $105\text{k}$ & $630\text{k}$  \\
  Warmup steps             & $1,000$   & $2,500$ & $675$  & $31,500$\\
  Volume loss weight       & $0.2$ & $0.2$  & $0.2$ & $0.2$  \\
\bottomrule
\end{tabular}
\end{table}

\paragraph{Parameter counts}
Table~\ref{tab:model_parameters} reports the number of trainable parameters of CANTO for each dataset, separating those associated with the geometry encoder from those of the remaining model components. The total parameter count ranges from 12.61M to 17.17M across datasets, with the variation stemming almost entirely from the geometry encoder (see Section~\ref{tokenizing_geometries}), whose size depends on the dataset-specific geometry configuration. In contrast, the remainder of the model architecture is shared across datasets and maintains an approximately constant parameter count of 8M parameters.

\begin{table}[ht]
\centering
\caption{\textbf{Model parameter counts across datasets.} Parameter counts are separated into geometry-related and non-geometry parameters.}
\label{tab:model_parameters}
\setlength{\tabcolsep}{8pt}
\renewcommand{\arraystretch}{1.15}
\begin{tabular}{@{}lrrr@{}}
\toprule
\textbf{Dataset} & \textbf{All} & \textbf{Geometry} & \textbf{Non-geometry} \\
\midrule
AhmedML      & 12.61M & 4.82M & 7.79M \\
WindsorML    & 13.47M & 5.46M & 8.01M \\
DrivAerML    & 17.17M &  9.24M & 7.93M \\
HiLiftAeroML & 13.47M &  5.46M & 8.01M \\
\bottomrule
\end{tabular}
\end{table}

\subsection{NURBS Anchor Split Resolution.}\label{appendix:anchor_split}
\begin{figure}[htbp]
  \centering
  \includegraphics[width=\linewidth]{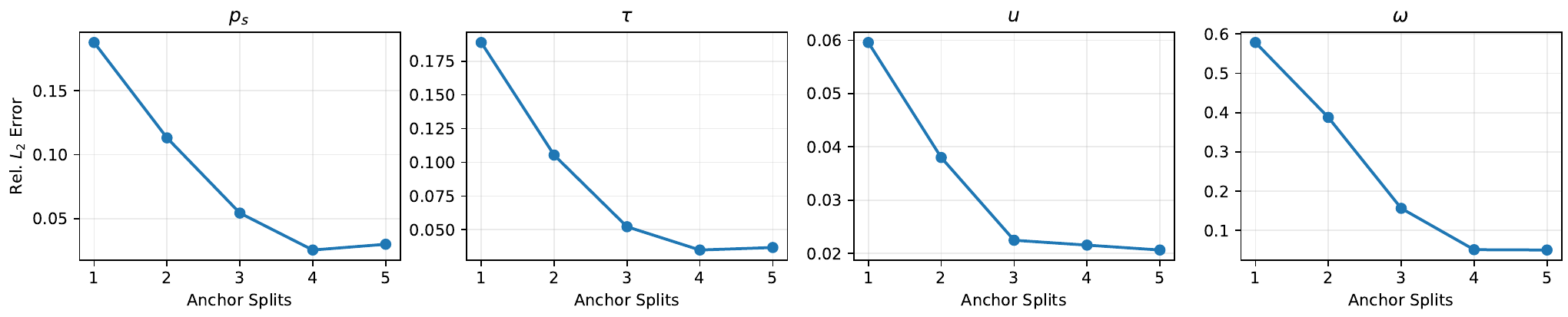}
  \caption{
  \textbf{Effect of the NURBS anchor split resolution on predictive accuracy.} Error drops sharply as the exact geometric context set is parametrically refined, plateauing at 4 splits, the level used for training.
  }
  \label{fig:anchor_split}
\end{figure}

Unlike point-cloud methods, \method generates surface anchors by parametrically splitting parent NURBS patches. Figure~\ref{fig:anchor_split} shows that increasing the number of anchor splits from 1 to 4 sharply reduces errors across all fields by providing a finer geometric context. Crucially, while \method was trained exclusively at 4 splits, it shows stable performance also at 5 splits. This suggests that the model learns a robust representation of the continuous geometry, remaining invariant to the specific parametric decomposition used during training.

\subsection{Point vs. Geometry Anchors.}\label{Pa-v-GA}

We consider another configuration of CANTO, closer to the AB-UPT backbone: \textbf{CANTO-PA} (``Point Anchors'') employs the NURBS-based geometry encoder described in Section~\ref{tokenizing_geometries} and samples surface anchors directly from the training point cloud.

\begin{figure}[htbp]
  \centering
  \includegraphics[width=\linewidth]{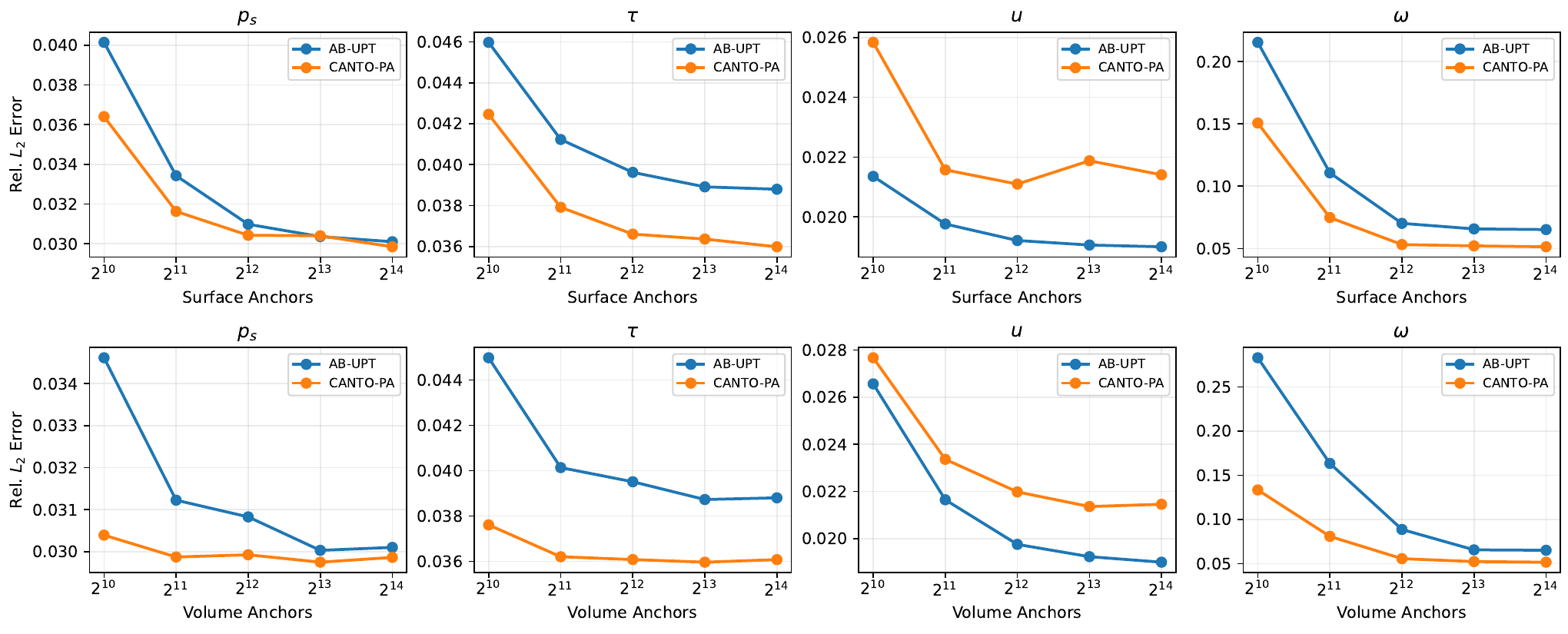}
  \caption{
   \textbf{Robustness to anchor sparsity for \textsc{CANTO-PA}.} Retaining point-based surface anchors causes performance to degrade when the number of surface anchors is reduced (a), even though the global CAD encoding provides strong robustness against volume sparsity (b).
  }
  \label{fig:canto_pa_sparsity}
\end{figure}

To isolate the benefit of the CAD encoder from the geometry anchors, Table~\ref{tab:windsorml:abl} compares \textsc{CANTO-PA} (which encodes the NURBS geometry but retains spatially sampled point anchors for the surface) against our primary \textsc{CANTO} model on WindsorML. Furthermore, Figure~\ref{fig:canto_pa_sparsity} illustrates that while the CAD encoder alone grants \textsc{CANTO-PA} remarkable robustness to volume sparsity (b), the model still degrades as the surface anchor count drops (a). This confirms that extracting surface tokens directly from exact parametric patches is essential to completely bypass spatial sampling limits.

\begin{table}[h]
\centering
\caption{
\textbf{Ablation of \method variants on the WindsorML dataset.}
}
\label{tab:windsorml:abl}
\setlength{\tabcolsep}{4pt}
\begin{tabular}{l ccc}
\toprule
\textbf{Method} & $\boldsymbol{p}_s$  & $\boldsymbol{u}$  & $\boldsymbol{\tau}$ \\
\midrule
\textsc{CANTO-PA} & 8.04 & 7.30 &  7.55 \\
\textsc{CANTO} & \textbf{6.81} & \textbf{6.69} & \textbf{5.46}   \\
\bottomrule
\end{tabular}
\end{table}

\subsection{Model robustness for varying number of anchors}\label{appendix:ablations}

\begin{figure}[h!]
  \vspace{-4mm}
  \centering
  \includegraphics[width=\linewidth]{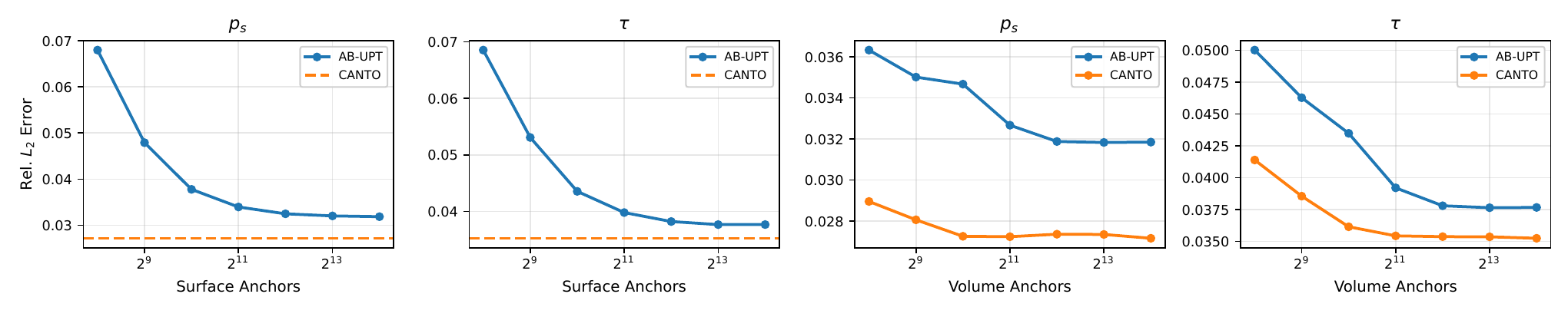}
  \caption{
   \textbf{Robustness to anchor sparsity for surface fields.} Relative L2 error is reported for pressure ($p_s$) and wall shear stress ($\tau$) against varying the number of surface anchors (left two panels) and volume anchors (right two panels). CANTO completely bypasses the need for spatial surface anchors (shown as a flat dashed line). Furthermore, it maintains high accuracy on these surface fields even at lower volume anchor counts where the AB-UPT baseline degrades sharply.
  }
  \label{fig:anchor_sparsity_main}
\end{figure}

We compare the anchor efficiency of \method against AB-UPT by varying the number of surface and volume anchors (Figure~\ref{fig:anchor_sparsity_main}). For point-cloud baselines like AB-UPT, reducing the number of surface anchors from $2^{14}$ to $2^{8}$ causes a catastrophic collapse in performance for surface fields (the error more than doubles for the surface pressure field). In contrast, because \method derives its surface tokens directly from the analytical CAD geometry, it completely bypasses spatial surface sampling. Its performance is therefore fully decoupled from the mesh resolution (represented by the flat dashed line) and outperforms the baseline even when AB-UPT uses the maximum $2^{14}$ surface anchors. 

\method's superior robustness also extends to the volume domain. As the number of volume anchors is reduced to a sparsity of $2^{8}$ (256 anchors), AB-UPT's error increases sharply. However, \method remains remarkably robust. Because the geometry-derived context set captures exact boundary conditions and continuous structural information, the model's physical predictions on the surface ($p_s$, $\tau$) rely far less on dense volume grids. This geometric foundation grants extreme sample efficiency, allowing \method to drastically reduce the attention context set ($M$) for substantial computational savings without sacrificing physical accuracy.

{
The previous section demonstrated the strong robustness of \method for predicting surface fields. We also evaluate the effect of anchor sparsity on the volumetric fields, $u$ and $\omega$ (Figure~\ref{fig:anchor_sparsity_volume}). Similar to the surface fields, \method completely bypasses the need for spatial surface anchors, maintaining a flat error curve for both volume velocity and vorticity as the number of surface anchors is reduced.
Furthermore, when reducing the number of volume anchors, \method still exhibits superior robustness compared to AB-UPT, delaying the performance degradation to much lower anchor counts.
\begin{figure}[h!]
  \vspace{-4mm}
  \centering
  \includegraphics[width=\linewidth]{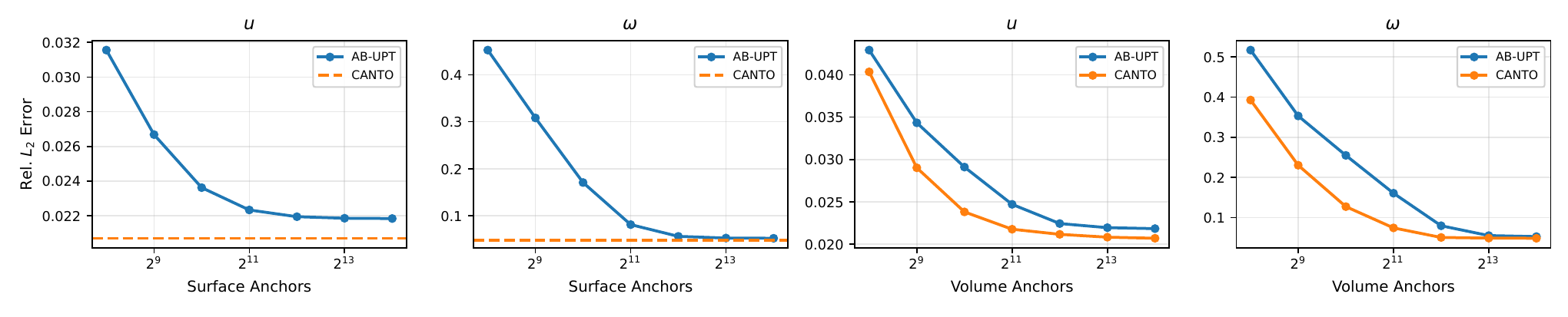}
  \caption{
   \textbf{Robustness to anchor sparsity for volume fields.} Relative L2 error is reported for volume velocity ($u$) and volume vorticity ($\omega$) against varying surface anchor counts (left two panels) and volume anchor counts (right two panels). \method's performance on volumetric fields is completely decoupled from the surface anchor sampling resolution (represented by the flat dashed line), and exhibits superior robustness to volume sparsity compared to AB-UPT.
  }
  \label{fig:anchor_sparsity_volume}
\end{figure}
}

\section{Model performance}\label{performance}

\subsection{Computational resources}\label{appendix:compute}

Table~\ref{tab:training_configs} reports the hardware, numerical precision, number of training epochs, and wall-clock training time used for each dataset. All experiments were run on a single node.

\begin{table}[ht]
\centering
\caption{Computational resources and training time per dataset. All runs use a single node; per-step throughput is dominated by attention over the volume-anchor set.}
\label{tab:training_configs}
\setlength{\tabcolsep}{10pt}
\renewcommand{\arraystretch}{1.15}
\begin{tabular}{@{}lcccc@{}}
\toprule
\textbf{Dataset} & \textbf{Hardware} & \textbf{Precision} & \textbf{Epochs} & \textbf{Training time} \\
\midrule
AhmedML   & $8\times$ H100  & FP32         & 1140 & 17\,h \\
WindsorML & $8\times$ H100  & FP32         & 1326 & 13\,h \\
DrivAerML & $4\times$ GB200 & FP16 (mixed) &  500 & 11\,h \\
HiLiftAeroML & $4\times$ GB200 & FP16 (mixed) &  500 & 69\,h \\
\bottomrule
\end{tabular}
\end{table}

% \subsection{Traditional Simulation Pipeline}

% \begin{figure*}[t]
%     \centering
%     \includegraphics[width=\textwidth]{Figures/traditional_pipeline.png}
%     \caption{
%     \textbf{Traditional CAD-to-simulation pipeline}.
%     Exact parametric CAD geometry is discretized into an adaptive mesh before numerical simulation, so both classical solvers and mesh-based learned surrogates operate on a sampled proxy rather than the native design representation.
%     }
%     \label{fig:traditional_pipeline}
% \end{figure*}

\subsection{Inference latency and peak memory requirement}\label{appendix:latency}

As shown in Table~\ref{tab:inference_latency}, our method scales to very large prediction problems, performing inference on up to 30 million surface points and 200 million volume points while maintaining a memory footprint below 55 GB. Notably, the geometry encoder (see Figure~\ref{fig:encoder}) uses the FlashAttention kernel, allowing us to deal with potentially a high number of token surfaces. Despite the substantial variation in CAD and mesh complexity across the datasets, inference remains practical, requiring approximately 75–160 seconds for the reported cases. The inference time represents the average inference time per element over the whole test set over 5 runs. The throughput is given as the number of query points inferred per second as indicated in Section~\ref{appendix:tradeoff}. 
Peak memory usage and inference time are directly linked through the chunk size of the query points processed during inference.

\begin{table}[ht]
\centering
\caption{Inference latency and peak memory.}
\label{tab:inference_latency}
\setlength{\tabcolsep}{4pt}
\renewcommand{\arraystretch}{1.15}
\begin{tabular}{@{}lcccccc@{}}
\toprule
\textbf{Dataset} & \textbf{Hardware} & \textbf{Peak mem} &
\textbf{Surf. pts} & \textbf{Vol. pts} &
\textbf{Inference time} & \textbf{Throughput} \\
\midrule
AhmedML      & H100  & 13\,GB & $\sim$1M  & $\sim$22M  & $\sim$75\,s  & $\sim$0.3M queries/s \\
WindsorML    & H100  & 55\,GB & $\sim$5M  & $\sim$290M & $\sim$330\,s & $\sim$0.9M queries/s \\
DrivAerML    & GB200 & 40\,GB & $\sim$9M  & $\sim$150M & $\sim$160\,s & $\sim$1.0M queries/s \\
HiLiftAeroML & GB200 & 42\,GB & $\sim$30M & $\sim$200M & $\sim$155\,s & $\sim$1.5M queries/s \\
\bottomrule
\end{tabular}
\end{table}

\subsection{Inference runtime-memory trade-off}\label{appendix:tradeoff}

Since CANTO predicts physical fields at arbitrary query locations, the query points are processed in chunks whose size can be adjusted according to the available GPU memory. Larger chunk sizes reduce the overall evaluation time at the cost of higher peak memory usage. The measured trade-off for CANTO on a full AhmedML evaluation on a single H100 GPU, for 22 million volume query points as well as 1 million surface query points, is summarized Table~\ref{tab:chunk_size}. These results demonstrate that inference can be adapted to the available GPU memory with only a moderate impact on runtime. For example, reducing the peak memory requirement from 22.7 GB to 3.6 GB increases the evaluation time from 131 s to 209 s using a chunk size of 50,000, while still remaining orders of magnitude faster than the underlying CFD simulation, as detailed in Section~\ref{pipeline_comparison}.

\begin{table}[t]
\centering
\caption{Effect of query chunk size on inference runtime and peak memory.}
\label{tab:chunk_size}
\setlength{\tabcolsep}{5pt}
\renewcommand{\arraystretch}{1.15}
\begin{tabular}{@{}lrrrrrr@{}}
\toprule
\textbf{Chunk size} & \textbf{10k} & \textbf{50k} & \textbf{100k} &
\textbf{250k} & \textbf{500k} & \textbf{1M} \\
\midrule
Eval. runtime & 543\,s & 209\,s & 169\,s & 146\,s & 138\,s & 131\,s \\
Peak memory   & 3.6\,GB & 3.6\,GB & 4.5\,GB & 7.6\,GB & 12.6\,GB & 22.7\,GB \\
\bottomrule
\end{tabular}
\end{table}

\subsection{End-to-end performance comparison with traditional CFD}\label{pipeline_comparison}

Table~\ref{tab:pipeline_speedup} compares the end-to-end runtime of our approach against the corresponding CFD workflows. AhmedML and DrivAerML were generated using OpenFOAM, WindsorML using Volcano ScaLES, and HiLiftAeroML using Fidelity CHARLES. For AhmedML and DrivAerML, where both CANTO and CFD timings are available, our model achieves orders-of-magnitude faster inference than the underlying CFD simulations: approximately 2,300× for AhmedML and 900× for DrivAerML compared with the CFD solve alone.

When the one-time CAD untrimming step is included in the complete raw-CAD-to-field workflow rather than treated as offline preprocessing, the wall-clock ratio remains approximately 2,300× for AhmedML, which is already untrimmed, and greater than 48× for DrivAerML, even without including the unavailable meshing time. Untrimming is intended as an offline preprocessing step and was designed for high dataset throughput rather than low per-case latency. The current implementation processes each geometry serially while processing multiple geometries concurrently across a 144-core CPU node. The reported ratios include the current untrimming implementation; further engineering optimization may reduce this preprocessing cost. Because CANTO and the CFD solvers run on different hardware, these values are wall-clock ratios rather than hardware-normalized speedups.

\begin{table}[ht]
\centering
\caption{\textbf{Inference and preprocessing cost compared with the CFD pipeline.}
Inference speedup compares model inference with the CFD solve, while pipeline
speedup compares untrimming and inference with CFD meshing and solve. ``--'' indicates that the corresponding information was not reported in the reference paper.}
\label{tab:pipeline_speedup}
\setlength{\tabcolsep}{3.5pt}
\renewcommand{\arraystretch}{1.15}
\begin{tabular}{@{}lccccccc@{}}
\toprule
\textbf{Dataset} & \textbf{GPU} & \textbf{Untrim.} &
\textbf{Inference} & \textbf{Meshing} & \textbf{CFD} &
\textbf{Inf. speedup} & \textbf{Pipeline speedup} \\
\midrule
AhmedML      & H100  & 0\,s          & $\sim$75\,s  & 30\,min & 48\,h & $\sim$2300$\times$ & $\sim$2300$\times$ \\
WindsorML    & H100  & $\sim$5\,s    & 330\,s        & 2\,min  & 28\,h & $\sim$300$\times$  & $\sim$300$\times$ \\
DrivAerML    & GB200  & $\sim$2800\,s & $\sim$160\,s & --      & 40\,h & $\sim$900$\times$  & $\sim$50$\times$ \\
HiLiftAeroML & GB200 & $\sim$1000\,s & $\sim$155\,s & --      & --    & --                   & -- \\
\bottomrule
\end{tabular}
\end{table}

\section{Datasets}\label{appendix:dataset}

\begin{figure}[t]
  \centering
  \includegraphics[width=0.3\linewidth]{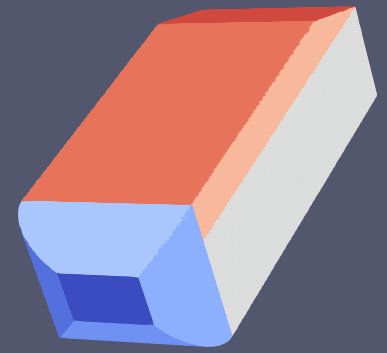}
  \caption{Surface decomposition of an AhmedML geometry into 13 untrimmed NURBS patches. The decomposition consists of nine planar surfaces and four ellipsoidal surfaces near the front, with each patch shown in a different color..}
  \label{ahmed_cad}
\end{figure}

\textbf{AhmedML.} The AhmedML dataset~\citep{ashton2024ahmedml} is an open-access dataset containing high-resolution computational fluid dynamics (CFD) simulations for 500 distinct geometric variants of the Ahmed body, a widely used benchmark in automotive aerodynamics. The simulations are performed in OpenFOAM using a hybrid RANS–LES turbulence modeling approach, capturing important flow phenomena such as pressure-induced separation and complex three-dimensional vortex structures. Each case is discretized on meshes comprising approximately 1 million surface cells and 20 million volume cells. The dataset was split into 400 training, 50 validation, and 50 test samples, following~\citep{alkin2025ab}. The exact split is publicly available at~\url{https://github.com/Emmi-AI/noether/blob/main/src/noether/data/datasets/cfd/caeml/ahmedml/split.py}. Notably, the AhmedML dataset does not distribute CAD files for each configuration. However, each AhmedML geometry is defined by explicit geometric design parameters. We use these parameters to reconstruct the corresponding analytic CAD geometry exactly as 13 untrimmed NURBS surfaces per case, as illustrated in Fig.~\ref{ahmed_cad}. This reconstruction is not obtained by fitting to a mesh or point cloud; it is generated directly from the parametric definition of the AhmedML design family. The resulting representation consists of nine planar surfaces and four curved surfaces near the front of the body.
Two versions of the NURBS surfaces were utilized in this paper. A first \textit{Variable-size} version parametrizes surfaces in the way a CAD file would, that is
\begin{itemize}
\item Planar surfaces are parametrized by NURBS surfaces of bi-degree $(1, 1)$, with four control points corresponding to the vertices of these surfaces and associated uniformly unit weights, and knot vectors $U = V = (0, 0, 1, 1)$
\item Curved surfaces parametrized by NURBS surfaces of bi-degree $(2, 2)$, with nine control points corresponding to the four vertices of these surfaces, the middle of the straight edges, three vertices at the intersection of the tangents in the previously constructed points. The associated weight matrix is given by
\[
W =
\begin{bmatrix}
1 & 1 & 1 \\
\frac{\sqrt{2}}{2} & \frac{\sqrt{2}}{2}  & \frac{\sqrt{2}}{2}  \\
1 & 1 & 1
\end{bmatrix}
\]
and knot vectors $U = V = (0, 0, 0, 1, 1, 1)$
\end{itemize}
A second \textit{Fixed-size} version uniformizes all surfaces, including the planar ones, to bi-quadratic surfaces with nine control points and knot vectors $U = V = (0, 0, 0, 1, 1, 1)$ (this is simply done by inserting, for the planar surfaces, control points at the middle of each edge and at the center of the surface).

\textbf{WindsorML.} WindsorML~\citep{ashton2024windsorml} is a large-scale, high-resolution dataset comprising 350 geometric configurations of the Windsor body. Each configuration is simulated using wall-modeled large-eddy simulation (WMLES) on computational grids exceeding 280 million cells, with a Cartesian immersed-boundary approach to ensure accurate aerodynamic predictions. The dataset provides parameterized CAD models in \texttt{.stp} format, containing 42 NURBS surfaces each, defined by seven geometric design variables relevant to automotive applications, along with aerodynamic force and moment coefficients. All cases are evaluated under identical flow conditions. As no explicit data split is provided in the original paper, we follow the proposed 60/20/20 partition using an ordered split without shuffling: the first 60\% of samples are used for training, the next 20\% for validation, and the final 20\% for testing.

\textbf{DrivaerML.}  The DrivaerML dataset~\citep{ashton2024drivaerml} comprises 500 parametrically modified variants of DrivAer vehicle geometries~\citep{heft2012experimental}, and was created to address the limited availability of open-access, large-scale computational fluid dynamics (CFD) data for automotive aerodynamics. The CFD simulations were performed on volumetric meshes containing approximately 140 million cells using a hybrid RANS–LES methodology Reynolds-Averaged Navier–Stokes–Large Eddy Simulation, which represents one of the highest-fidelity approaches currently employed in the automotive industry. Each sample includes surface meshes with roughly 8.8 million points, providing surface pressure and wall shear stress data, as well as volumetric fields for velocity, pressure, and vorticity. The available data, which includes the 500 variants minus 16 unavailable geometries, are divided into 400 training samples, 34 validation samples, and 50 test samples, following the split used in~\citep{alkin2025ab}, and provided at~\url{https://github.com/Emmi-AI/noether/blob/main/src/noether/data/datasets/cfd/caeml/drivaerml/split.py}. CAD geometries under .stp format were provided to us, each containing roughly 8.5k NURBS surfaces.

\textbf{HiLiftAeroML.}   HiLiftAeroML~\citep{ashton2026hiliftaeromlhighfidelitycomputationalfluid} is a large-scale, high-fidelity CFD dataset for high-lift aircraft aerodynamics, comprising 1,800 simulations generated from 180 parametric variants of the NASA Common Research Model in high-lift configuration (CRM-HL), each evaluated at 10 angles of attack ranging from $4^\circ$ to $22^\circ$ in $2^\circ$ increments. The geometries span an eight-dimensional design space controlling the inboard and outboard slat and flap deflection angles and gap multipliers, covering a broad range of high-lift configurations. Simulations were performed using the GPU-accelerated Fidelity Charles solver with explicit wall-modeled large-eddy simulation (WMLES) on solution-adapted Voronoi grids containing approximately 300--500 million control volumes, enabling the resolution of complex separated flows and both pre- and post-stall regimes. Each sample provides the aircraft geometry, time-averaged surface and volumetric flow fields, as well as integrated aerodynamic forces and moments. We use the full in-distribution split proposed by the authors, consisting of 1,260 training, 180 validation, and 360 test cases, corresponding to a 70/10/20 partition of the complete dataset. The angle of attack is incorporated by rotating the input CAD geometry and corresponding mesh before passing them to \method.

\section{Supplemental NURBS background and untrimming}\label{app:NURBS}

\subsection{B-spline basis functions}\label{app:nurbs_param}

Given a knot vector $U$, the degree-$p$ B-spline basis functions $N^U_{k, p}$ ($k \geq 0$) are defined recursively.
For $p=0$,
\[
  N_{k,0}^{U} (u) =
  \begin{cases}
    1, & \text{if } u_k \le u < u_{k+1}, \\
    0, & \text{otherwise},
  \end{cases}
\]
and for $p \ge 1$,
\[
    N_{k,p}^U (u)
  = \alpha_{k,p}(u) \, N^U_{k,p-1}(u)
  + \beta_{k,p}(u) \, N^U_{k+1,p-1}(u),
\]
where
\[
    \alpha_{k,p}(u) = 
    \begin{cases}
        \frac{u - u_k}{u_{k+p} - u_k}, & u_{k+p} \neq u_k, \\
        0, & \text{otherwise}
    \end{cases}, \quad
    \beta_{k,p}(u) = 
    \begin{cases}
        \frac{u_{k+p+1} - u}{u_{k+p+1} - u_{k+1}}, & u_{k+p+1} \neq u_{k+1}, \\
        0, & \text{otherwise}.
    \end{cases}
\]

\subsection{Untrimming} % Boundary-First Untrimming
\label{app:untrimming}

Working with the trimming curves directly can be a practical challenge due to the complex nature of the way they specify the trimmed parameter domain $\mathcal{D}$. The work of~\citep{massarwi2018untrimming} proposes algorithms to untrim a NURBS surface by decomposing it into a collection of tensor-product NURBS patches. For a given patch $\Sigma$, this yields a decomposition $\Sigma = \bigcup_{j=1}^{n_{\Sigma}} \Sigma_{j},$
where each $\Sigma_j$ is diffeomorphic to $[0,1]^2$ and represented by a NURBS parameterization $S_{j}$ given by \eqref{eq:parametric_surface}. We denote by $n_{\Sigma}$ the number of untrimmed surfaces and $S_{j}$ their respective maps for $k=1,\dots,n_{\Sigma}$. 
These maps form the geometric inputs to our model (Section~\ref{tokenizing_manifolds}).

While exact, the untrimming procedure of~\citep{massarwi2018untrimming} significantly increases the number of patches, the bi-degree of the NURBS surfaces, and the number of control points. 
We instead use a boundary-first construction (Appendix~\ref{app:untrimming}) that exactly preserves watertightness, reproduces the input surface up to a small geometric residual (Table~\ref{tab:surface-error}), and keeps the number of patches, the bi-degree, and the number of control points comparable to those of the input.

\paragraph{Construction}

Given a trimmed patch $\Sigma$, let $S : [0,1]^2 \to \mathbb{R}^3$ be its underlying NURBS mapping of bi-degree $(p,q)$. The trimmed parameter domain $\mathcal{D} \subseteq [0,1]^2$ is bounded by closed loops of NURBS edge curves $\mathcal{T} = {\mathbf{c}^{\ell}}$.

We decompose $\Sigma$ into a sequence of untrimmed rational tensor-product NURBS patches $\Sigma_k$, each diffeomorphic to $[0,1]^2$ and parameterized by

\[
    S_k : [0,1]^2 \to \mathbb{R}^3.
\]

Each $S_k$ is defined by knot vectors $(U^{(k)},V^{(k)})$ and a weighted control net $(P^{(k)}_{ij},w^{(k)}_{ij})$.

As detailed below, the four boundary isocurves of each patch, obtained by fixing one parametric coordinate to $0$ or $1$, are exact pieces of the input trim. Watertightness therefore holds by construction: adjacent patches use a common lift of their shared edge and share the corresponding row of control points.

The trimmed parameter domain $\mathcal{D}$ of patch $\Sigma$ is decomposed via the ear-clipping algorithm~\citep{meisters1975polygons} into four-edge subdomains $\mathcal{D}_m$ bounded by canonical edge curves $(\mathbf{c}_m^{u_0}, \mathbf{c}_m^{u_1}, \mathbf{c}_m^{v_0}, \mathbf{c}_m^{v_1})$. The four curves are lifted to their least common refinement by knot insertion and degree elevation, then assigned as the four boundary isocurves of $\Sigma$. The interior control points are determined by collocating the patch at Greville abscissae to match the parent surface evaluated through a transfinite Coons blend. Writing $\boldsymbol{\varphi}_m : [0, 1]^2 \to \Sigma_m$ for the Coons blend that interpolates the four edge curves in the parent's UV and $(\xi_i, \eta_j)$ for the Greville abscissae of $(\bar{U}_m, \bar{V}_m)$, the interior weighted control points solve
\[
  S_m(\xi_i, \eta_j) \;=\; S\bigl(\boldsymbol{\varphi}_m(\xi_i, \eta_j)\bigr)
\]
at every interior node with boundary control points held fixed at their lifted values. The system is solved by Tikhonov-regularised least squares in the rank-deficient case. When $\boldsymbol{\varphi}_m$ is affine, $\mathbf{S}_k \circ \boldsymbol{\varphi}_m$ is itself a tensor-product NURBS and the construction is exact.

The control points of all emitted patches form a single homogeneous pool $\mathbf{P}_h \in \mathbb{R}^{N \times 4}$ whose rows store $(w_{ij}\mathbf{P}_{ij}, w_{ij})$. Patch $\Sigma_m$ is specified by its knot vectors $(U^{(m)}, V^{(m)})$. Two patches sharing a canonical trim edge have identical pool indices on the corresponding boundary row, so watertightness is encoded in the index structure alone and holds across heterogeneous parent surfaces.

\paragraph{Empirical evaluation}

We report per-patch surface preservation error in Table~\ref{tab:surface-error}: the worst case is sub-millimetre and the median is several orders of magnitude smaller.

\begin{table}[ht]
\centering
\caption{Per-patch surface preservation error of the untrimming pipeline on the Windsor dataset. Per-patch error is the maximum, over an $8 \times 8$ uniform grid in $[0, 1]^2$, of the Newton-projected perpendicular distance from the output sample to the original CAD parent surface; we report the maximum and median across patches, normalised by the model's 3-D bounding-box diagonal.}
\label{tab:surface-error}
\begin{tabular}{lcrr}
\toprule
Dataset & Bbox diag.\ (m) & Max rel.\ error & Median rel.\ error \\
\midrule
Windsor & $1.19$ & $6.40 \times 10^{-4}$ & $1.73 \times 10^{-14}$ \\
\bottomrule
\end{tabular}
\end{table}

\paragraph{Untrimming memory requirements and runtime}\label{untrimming_runtime}

The current untrimming implementation processes each geometry serially on the CPU. At the dataset level, multiple geometries are processed concurrently on a 144-core CPU node with 2 TB of RAM. The implementation was designed for high dataset-level throughput rather than minimum latency for an individual geometry. Table~\ref{tab:untrimming_runtime} reports both the per-geometry measurements and the time required to preprocess each complete dataset.

Untrimming time varies with CAD complexity, including the number of trimmed surfaces, spline degrees, control points, trimming curves, and holes. This preprocessing is performed once per geometry, and the resulting parametric NURBS representation can be reused for subsequent model evaluations. It is separate from CANTO’s model-inference time.

\begin{table}[ht]
\centering
\caption{Untrimming computational cost.}
\label{tab:untrimming_runtime}
\setlength{\tabcolsep}{4pt}
\renewcommand{\arraystretch}{1.15}
\begin{tabular}{@{}lcccrr@{}}
\toprule
\textbf{Dataset} & \textbf{Runtime} & \textbf{Memory} &
\textbf{Full preprocessing} & \textbf{Initial surf.} & \textbf{Final surf.} \\
\midrule
WindsorML    & $\sim$5\,s    & 0.5\,GB & $\sim$15\,s   & 42       & 71 \\
DrivAerML    & $\sim$2800\,s & 3.9\,GB & $\sim$8\,h    & $\sim$8.5k & $\sim$45k \\
HiLiftAeroML & $\sim$1000\,s & 2.3\,GB & $\sim$10\,h  & $\sim$350  & $\sim$4.5k \\
\bottomrule
\end{tabular}
\end{table}

\section{Inverse design and Optimization}\label{Optimization}

\paragraph{Problem statement.}
We minimize aerodynamic drag over the six-parameter AhmedML body family subject
to a lower bound on enclosed volume and an upper bound on lift, as expressed in the problem~\eqref{eq:problem}. $D$ and $L$ are the drag and lift coefficients evaluated at the
\emph{fixed} dataset reference area $A_{\mathrm{ref}} = 0.112032\,\mathrm{m^2}$
(the nominal Ahmed cross-section $0.389 \times 0.288\,\mathrm{m}$), so they are
proportional to drag and lift \emph{force} rather than being normalized per
geometry. The volume $V$, in turn, is the enclosed volume of the differentiable
geometry map, obtained in closed form from the NURBS control net. The value $0.05$ for the lift constraint defines a non-degenerate feasible set, comprising $298$ of the $500$ designs in the dataset.

\paragraph{Design space.}
The six parameters and their dataset bounds $[x_i^{\mathrm{lo}}, x_i^{\mathrm{hi}}]$
are given in Table~\ref{tab:design-space}, and define the feasible box
$\mathcal{X} = \{x \in \mathbb{R}^6 : x^{\mathrm{lo}} \leq x \leq x^{\mathrm{hi}}\}$.
The optimization is carried out in normalized coordinates obtained by the
affine map
\begin{equation}
\label{eq:normalize}
z_i = \frac{x_i - x_i^{\mathrm{lo}}}{s_i},
\qquad
x_i = x_i^{\mathrm{lo}} + s_i z_i ,
\qquad
s_i = x_i^{\mathrm{hi}} - x_i^{\mathrm{lo}} ,
\qquad i = 1,\dots,6 ,
\end{equation}
which carries $\mathcal{X}$ onto the unit cube $[0,1]^6$, so the bound
constraints become $0 \leq z \leq 1$ and every objective and constraint
gradient transforms by the diagonal scaling.

\begin{table}[ht]
\vspace{-10pt}
\centering
\captionsetup{skip=1pt}
\caption{Design parameters and bounds (mm).}
\label{tab:design-space}
\begin{tabular}{lrr}
\toprule
Parameter & Lower & Upper \\
\midrule
body length          & 800.0 & 1200.0 \\
body height          & 250.0 &  315.0 \\
body width           & 300.0 &  500.0 \\
front arc diameter   & 160.0 &  240.0 \\
slant angle length   &  75.0 &  246.2 \\
slant angle height   &  26.0 &  216.5 \\
\bottomrule
\end{tabular}
\end{table}

\paragraph{Drag and Lift integration}

Drag and lift are computed by integrating the surrogate-predicted surface coefficients directly over the CAD geometry, without introducing a CFD mesh into the optimization loop. Each of the $13$ NURBS patches of the Ahmed body is evaluated on a $32\times32$ midpoint grid in parameter space, giving $13{,}312$ quadrature points in total. At each point, the exact NURBS derivatives provide the oriented normalized surface normal $\nu$ (see~\eqref{eq:surface_normal}) 
and the surface Jacobian
$\left|\frac{\partial S}{\partial u}\times\frac{\partial S}{\partial v}\right|$.
Thus, for any surface quantity $f$, we approximate its integral over a patch $\Sigma$ as

$$
\int_{\Sigma} f\,dA
\;\approx\;
\sum_{a=1}^{32}\sum_{b=1}^{32}
f(u_a,v_b)
\left\|
\frac{\partial S}{\partial u}(u_a,v_b)
\times
\frac{\partial S}{\partial v}(u_a,v_b)
\right\|
\Delta u\,\Delta v,
$$

where $(u_a,v_b)$ are the midpoint quadrature points.

At the same quadrature points, CANTO predicts the pressure and wall-shear stress $(p_s ,\tau)$, which are integrated to obtain the drag $D$ and lift $L$. We use the dataset's fixed reference area $A_{\mathrm{ref}}=0.112,\mathrm{m}^2$ for all designs.

\paragraph{Surrogate-based optimization.}
As indicated above, $D$ and $L$ are provided by a neural surrogate trained on the AhmedML dataset, which comprises $500$ CFD simulations: $400$ are used for training, while the remaining $100$ are held out for validation and testing. Since both the mapping from the design variables $\mathbf{z}$ to the NURBS geometry and the neural surrogate are differentiable, the gradients $\nabla_{\mathbf{z}} D$, $\nabla_{\mathbf{z}} V$, and $\nabla_{\mathbf{z}} L$ can be computed directly through automatic differentiation. We therefore solve~\eqref{eq:problem} using SLSQP~\citep{kraft1994algorithm}, providing these exact gradients to the optimizer. The optimization is initialized from the lowest-drag feasible design in the dataset and subsequently explores the continuous design space using the surrogate-derived gradients.

\paragraph{CFD validation.}
To validate the optimized designs, each geometry is tessellated at the same mesh density used in the original dataset and re-simulated following the full AhmedML CFD protocol described in~\citep{ashton2024ahmedml}. Specifically, the simulations are performed using OpenFOAM v2306, with \texttt{snappyHexMesh} used to generate meshes containing approximately $22$ million cells. The flow is then simulated using \texttt{pimpleFoam} for $133{,}333$ time steps, corresponding to a final simulation time of $t=80$,s. Consistent with the dataset protocol, the aerodynamic coefficients are time-averaged over the interval $t\in[20,80]$,s.

\paragraph{Results.}
Figure~\ref{fig:sweep_front_validated}  describes the validated front. Every optimized design
improves on the best dataset design at its volume, by $4.4\%$ to $20.5\%$, and
the surrogate's predictions are accurate to within $2.2\%$ throughout. The lift constraint is satisfied in
CFD at every threshold.

The optimized designs achieve gains across the entire front, with improvements ranging from $4.4\%$ to $20.4\%$ relative to the best dataset designs at the corresponding volume thresholds. At $V_0 = 70$ and $80$\,L, the volume constraint is inactive, and both thresholds converge to the same design with a volume of $84.3$\,L, corresponding to the unconstrained optimum within the considered design family. Table~\ref{tab:results} reports the results of these optimization and validation procedures in detail.

\begin{table}[h!]
% \begin{wraptable}{r}{0.3\textwidth}
\centering
\caption{\label{tab:results}
\textbf{CFD-validated optimized front}. $D_{\mathrm{cfd}}$ is the
protocol-averaged coefficient; ``dataset'' is the lowest-drag dataset design
satisfying the same constraints.}
\begin{tabular}{rcccccc}
\toprule
$V_0$ [L] & $L_\mathrm{cfd}$ & $D_{\mathrm{surr}}$ & $D_{\mathrm{cfd}}$ & error & dataset & gain \\
\midrule
 70 & 0.044 & 0.1566 & 0.1586 & $+1.3\%$ & 0.1867 & $+15.1\%$ \\
 80 & 0.044 & 0.1566 & 0.1586 & $+1.3\%$ & 0.1867 & $+15.1\%$ \\
 90 & 0.020 & 0.1649 & 0.1685 & $+2.2\%$ & 0.1867 & $+9.7\%$  \\
100 & 0.027 & 0.1811 & 0.1817 & $+0.4\%$ & 0.2285 & $+20.4\%$ \\
110 & 0.013 & 0.2004 & 0.2026 & $+1.1\%$ & 0.2317 & $+12.5\%$ \\
120 & 0.020 & 0.2196 & 0.2241 & $+2.1\%$ & 0.2589 & $+13.4\%$ \\
130 & -0.045 & 0.2393 & 0.2371 & $-1.0\%$ & 0.2656 & $+10.7\%$ \\
140 & 0.013 & 0.2586 & 0.2637 & $+2.0\%$ & 0.2971 & $+11.2\%$ \\
150 & 0.028 & 0.2790 & 0.2840 & $+1.8\%$ & 0.2971 & $+4.4\%$  \\
\bottomrule
\end{tabular}
\end{table}

\section{Qualitative Surface-Pressure Results}

\begin{figure}[htbp]
  \centering
  \includegraphics[width=\linewidth]{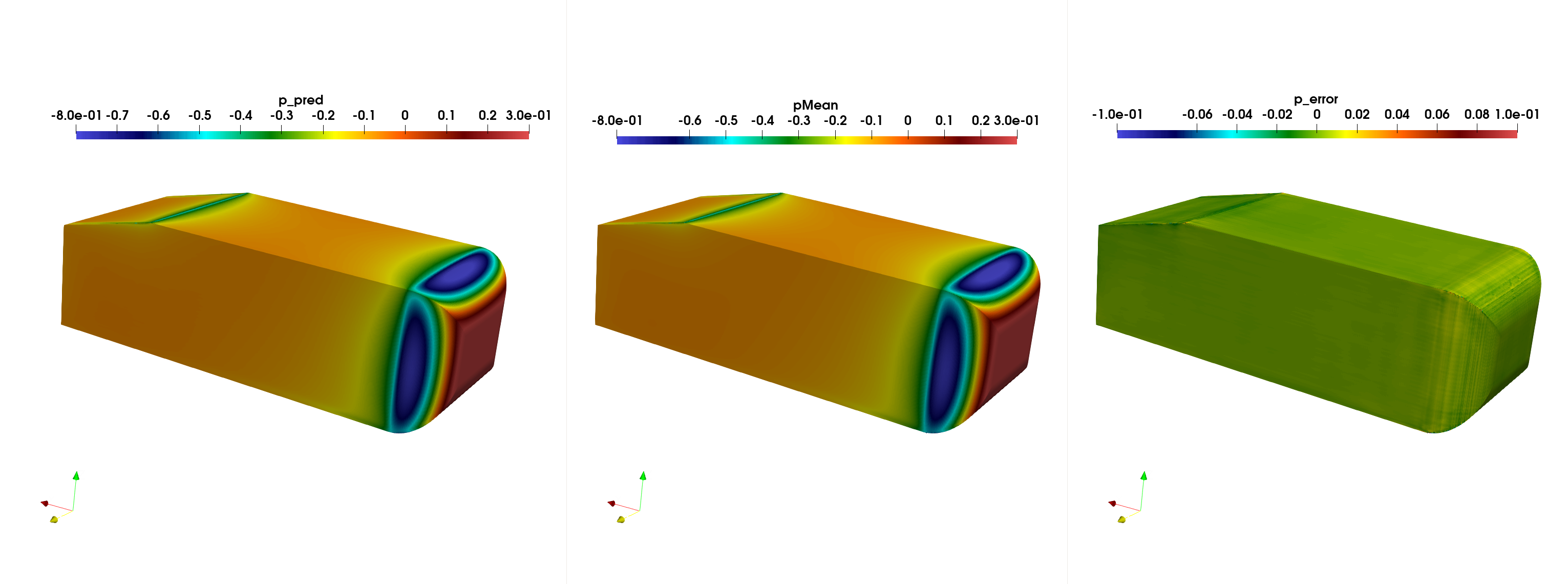}
  \caption{
  \textbf{Qualitative AhmedML surface-pressure prediction.} Left: \method prediction. Center: CFD reference field. Right: pointwise prediction error on the vehicle surface.
  }
  \label{fig:ahmed_qualitative_pressure}
\end{figure}

\begin{figure}[htbp]
  \centering
  \includegraphics[width=\linewidth]{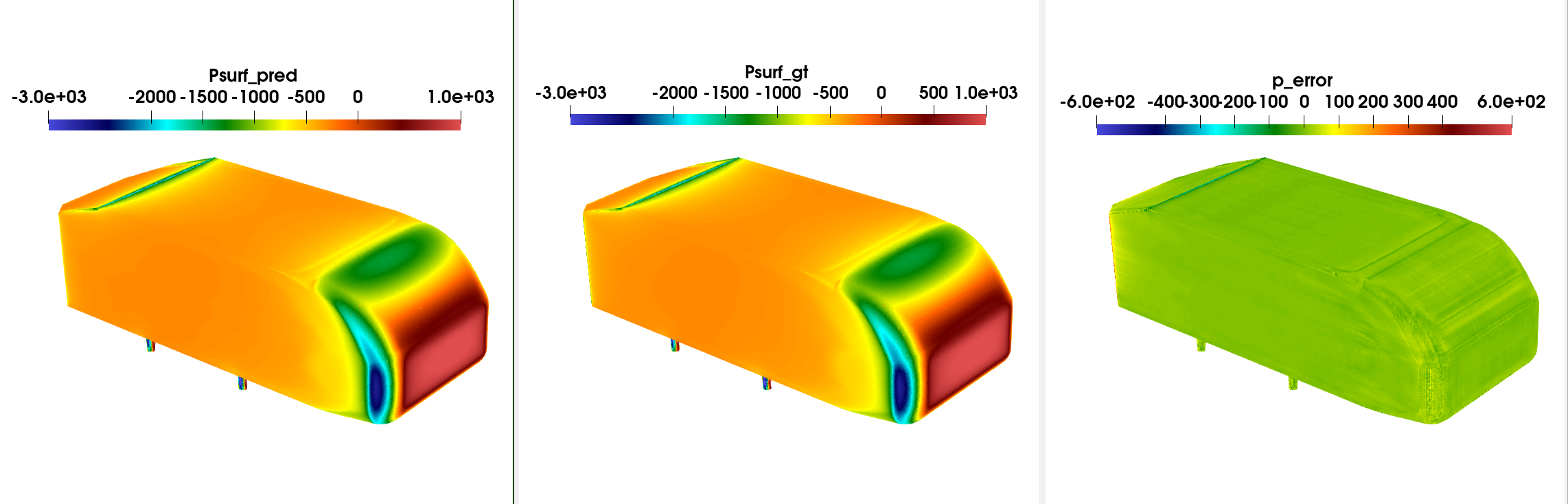}
  \caption{
  \textbf{Qualitative WindsorML surface-pressure prediction.} Left: \method prediction. Center: CFD reference field. Right: pointwise prediction error on the vehicle surface.
  }
  \label{fig:windsor_qualitative_pressure}
\end{figure}

Figures~\ref{fig:ahmed_qualitative_pressure} and~\ref{fig:windsor_qualitative_pressure} provide representative qualitative visualizations of surface-pressure predictions on AhmedML and WindsorML. In both cases, the predicted fields capture the large-scale pressure distribution and the dominant high-gradient structures, with the largest residuals concentrated near geometric transition regions.

\end{document}